\documentclass{article}

\usepackage{arxiv}

\usepackage{graphicx}       
\graphicspath{ {./} }       
\usepackage{amsmath}        
\usepackage{amssymb}        
\usepackage[utf8]{inputenc} 
\usepackage[T1]{fontenc}    
\usepackage{url}            
\usepackage{booktabs}       
\usepackage{amsfonts}       
\usepackage{nicefrac}       
\usepackage{microtype}      
\usepackage{todonotes}      
\usepackage{algorithm}      
\usepackage{algorithmic}    
\usepackage{hyperref}       
\usepackage{enumitem}       
\usepackage{mathtools}      
\usepackage{natbib}         
\setcitestyle{authoryear, round, citesep={;}, aysep={,}, yysep={;}}
\renewcommand{\cite}[1]{\citep{#1}}

\title{Multi-Task Learning with Deep Neural Networks: A Survey}
\date{}
\author{
  Michael Crawshaw \\
  Department of Computer Science \\
  George Mason University \\
  \texttt{mcrawsha@gmu.edu} \\
}

\renewcommand{\headeright}{}
\renewcommand{\undertitle}{}

\setlist[itemize]{leftmargin=*}

\begin{document}
\maketitle

\begin{abstract}
Multi-task learning (MTL) is a subfield of machine learning in which multiple tasks are
simultaneously learned by a shared model. Such approaches offer advantages like improved
data efficiency, reduced overfitting through shared representations, and fast learning
by leveraging auxiliary information. However, the simultaneous learning of multiple
tasks presents new design and optimization challenges, and choosing which tasks should
be learned jointly is in itself a non-trivial problem. In this survey, we give an
overview of multi-task learning methods for deep neural networks, with the aim of
summarizing both the well-established and most recent directions within the field. Our
discussion is structured according to a partition of the existing deep MTL techniques
into three groups: architectures, optimization methods, and task relationship learning.
We also provide a summary of common multi-task benchmarks.
\end{abstract}

\section{Introduction} \label{intro}
Multi-task learning is a training paradigm in which machine learning models are trained
with data from multiple tasks simultaneously, using shared representations to learn the
common ideas between a collection of related tasks. These shared representations
increase data efficiency and can potentially yield faster learning speed for related or
downstream tasks, helping to alleviate the well-known weaknesses of deep learning:
large-scale data requirements and computational demand. However, achieving such effects
has not proven easy and is an active area of research today.

We believe that MTL reflects the learning process of human beings more accurately than
single task learning in that integrating knowledge across domains is a central tenant of
human intelligence. When a newborn baby learns to walk or use its hands, it accumulates
general motor skills which rely on abstract notions of balance and intuitive physics.
Once these motor skills and abstract concepts are learned, they can be reused and
augmented for more complex tasks later in life, such as riding a bike or tightrope
walking. Any time that a human attempts to learn something new, we bring a tremendous
amount of prior knowledge to the table. It's no wonder that neural networks require such
numerous training examples and computation time: every task is learned from scratch.
Imagine trying to learn to tightrope walk without first learning to walk! The human
ability to rapidly learn with few examples is dependent on this process of learning
concepts which are generalizable across multiple settings and leveraging these concepts
for fast learning; we believe that developing systems to perform this process should be
the goal of multi-task learning and the related fields of meta-learning
\citep{hospedales_2020}, transfer learning \citep{zhuang_2019}, and continuous/lifelong
learning \citep{parisi_2019}.

Learning concepts for multiple tasks does bring difficulties which aren't present in
single task learning. In particular, it may be the case that different tasks have
conflicting needs. In this case, increasing the performance of a model on one task will
hurt performance on a task with different needs, a phenomenon referred to as
\textit{negative transfer} or \textit{destructive interference}. Minimizing negative
transfer is a key goal for MTL methods. Many architectures are designed with specific
features to decrease negative transfer, such as task-specific feature spaces and
attention mechanisms, but division of information between tasks is a fine line to walk:
we want to allow information flow between tasks that yields positive transfer, and
discourage sharing when it would create negative transfer. The question of how exactly
to design such a system is being actively investigated in MTL research.

The existing methods of MTL have often been partitioned into two groups with a familiar
dichotomy: \textit{hard parameter sharing} vs. \textit{soft parameter sharing}. Hard
parameter sharing is the practice of sharing model weights between multiple tasks, so
that each weight is trained to jointly minimize multiple loss functions. Under soft
parameter sharing, different tasks have individual task-specific models with separate
weights, but the distance between the model parameters of different tasks is added to
the joint objective function. Though there is no explicit parameter sharing, there is an
incentive for the task-specific models to have similar parameters. This is a useful
dichotomy, but the nature of multi-task methods has grown extremely diverse in the past
few years, and we feel that these two categories alone are not broad enough to
accurately describe the entire field. Instead, we widen the scope of the members of this
dichotomy to cover more ground. We generalize the class of hard parameter sharing
methods to multi-task architectures, while soft parameter sharing is broadened into
multi-task optimization. When combined, architecture design and optimization techniques
provide a nearly complete image of modern MTL. However, there is still an important
direction within the field that is missing even from this generalized dichotomy: task
relationship learning. Task relationship laerning (or TRL) methods focus on learning an
explicit representation of the relationships between tasks, such as task embeddings or
transfer learning affinities, and these types of methods don't quite fit into either
architecture design or optimization. Broadly speaking, these three directions -
architecture design, optimization, and task relationship learning - make up the existing
methods of modern deep multi-task learning.

Many different researchers have used the term multi-task learning to refer to different
settings, and we feel that it is important to clarify the scope of this review. As a
convention, we interpret MTL to only contain learning settings in which a fixed set of
tasks is learned simultaneously, and each task is treated equally. This means that we
don't consider training settings that have only a single ``main task" with one or more
auxiliary tasks, as well as settings in which the set of tasks to learn changes over
time. We may, however, discuss models which were designed for such settings, if the
ideas from the model are easily applicable to MTL.

The rest of the survey is outlined as follows. Section \ref{architectures} contains a
discussion of neural network architectures for multi-task learning. In section
\ref{optimization}, we discuss MTL optimization strategies, and we discuss methods for
learning explicit task relationships in section \ref{relationship}. Section
\ref{benchmarks} contains an overview of common multi-task benchmark for various
domains. Finally, we conclude with section \ref{conclusion}. Within each subsection or
subsubsection, the methods are mostly presented in order of publication, from earliest
to most recent. It should be noted that we do not discuss any classical (non-neural)
multi-task learning methods, though a thorough review can be found in \cite{zhang_2017,
ruder_2017}.

\section{Multi-Task Architectures} \label{architectures}
A large portion of the MTL literature is devoted to the design of multi-task neural
network architectures. There are many different factors to consider when creating a
shared architecture, such as the portion of the model's parameters that will be shared
between tasks, and how to parameterize and combine task-specific and shared modules.
More variations arise when considering architectures for a specific problem domain, like
how to partition convolutional filters into shared and task-specific groups for a set of
vision tasks. Many of the proposed architectures for MTL play a balancing game with the
degree of information sharing between tasks: Too much sharing will lead to negative
transfer and can cause worse performance of joint multi-task models than individual
models for each task, while too little sharing doesn't allow the model to effectively
leverage information between tasks. The best performing architectures for MTL are those
which balance sharing well. 

We partition the MTL architectures into four groups: architectures for a particular task
domain, multi-modal architectures, learned architectures, and conditional architectures.
For single-domain architectures, we consider the domains of computer vision, natural
language processing, and reinforcement learning. Multi-modal architectures handle tasks
with input in more than one mode, such as visual question answering with both a visual
and a language component. It should be noted that we only consider multi-modal
architectures which handle multiple tasks. For a more complete discussion of multi-modal
methods, see \cite{baltrusaitis_2019}. Lastly, We make the following distinction between
learned architectures and conditional architectures: Learned architectures are fixed
between steps of architecture learning, so the same computation is performed for each
input from the same task. In conditional architectures, the architecture used for a
given piece of data is dependent on the data itself.

\subsection{Architectures for Computer Vision} \label{cv_architectures}
In the single-task setting, many major developments for computer vision architectures
have focused on novel network components and connections to improve optimization and
extract more meaningful features, such as batch normalization \cite{ioffe_2015},
residual networks \cite{he_2016}, and squeeze and excitation blocks \cite{hu_2018}. In
contrast, many multi-task architectures for computer vision focus on partitioning the
network into task-specific and shared components in a way that allows for generalization
through sharing and information flow between tasks, while minimizing negative transfer.

\subsubsection{Shared Trunk} \label{shared_trunk} 
\begin{figure}[tb]
    \centerline{\includegraphics[scale=0.19]{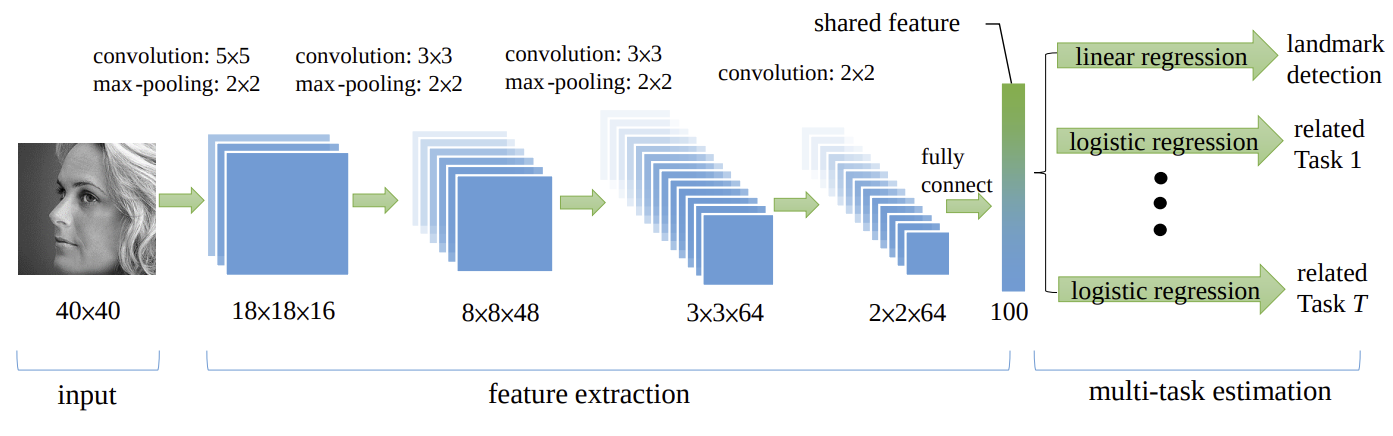}}
    \caption{Architecture for TCDCN \cite{zhang_2014}. The base feature extractor is
    made of a series of convolutional layers which are shared between all tasks, and the
    extracted features are used as input to task-specific output heads.}
    \label{tcdcn}
\end{figure}

Traditionally, many multi-task architectures in computer vision follow a simple outline:
A global feature extractor made of convolutional layers shared by all tasks followed by
an individual output branch for each task, as in figure \ref{tcdcn}. We will refer to
this template as a \textit{shared trunk}.

\cite{zhang_2014, dai_2016, zhao_2018, liu_2019a, ma_2018} propose architectures which
are variations on the shared trunk idea. \cite{zhang_2014}, the earliest of these works,
introduces Tasks-Constrained Deep Convolutional Network (TCDCN), whose architecture is
shown in figure \ref{tcdcn}. The authors propose to improve performance on a facial
landmark detection task by jointly learning head pose estimation and facial attribute
inference.  \cite{dai_2016} introduces Multi-task Network Cascades (MNCs). The
architecture of MNCs is similar to TCDCN, with an important difference: the output of
each task-specific branch is appended to the input of the next task-specific branch,
forming the ``cascade" of information flow after which the method is named. This type of
architecture is similar to the cascaded information networks for NLP discussed in
section \ref{cascade}.

\begin{figure}[tb]
    \centerline{\includegraphics[scale=0.25]{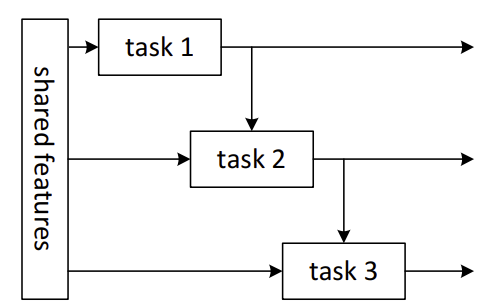}}
    \caption{Illustration of Multi-task Network Cascades \cite{dai_2016}. The output of
    the first task is used as an input for the second task, the second task's output is
    used as an input for the third task, and so on.}
    \label{mncs}
\end{figure}

\cite{zhao_2018, liu_2019a} each build on this original template with the introduction
of task-specific modules which can be placed within existing shared architectures. By
doing this, the computation of features relies on both the shared parameters of the
feature extractor and the task-specific parameters of modules placed through the
network, so that features of different tasks may differ before the task-specific output
branches. \cite{zhao_2018} introduces a modulation module in the form of a task-specific
channel-wise linear projection of feature maps, and the authors design a convolutional
architecture with these modules following convolutional layers in the latter half of the
network. Interestingly, it is empirically shown that the inclusion of these
task-specific projection modules decreases the chance that gradient update directions
for different tasks point in opposite directions, implying that this architecture
decreases the occurence of negative transfer. \cite{liu_2019a} proposes task-specific
attention modules. Each attention module takes as input the features from some
intermediate layer of the shared network concatenated with the output of the previous
attention module, if one exists. Each module computes an attention map by passing its
input through a Conv-BN-ReLU layer followed by a Conv-BN-Sigmoid layer. The attention
map is then element-wise multiplied with features from a successive shared layer, and
this product is the output of the attention module. This attention module allows the
network to emphasize features in the network which are more important for the
corresponding task, and downplay the effect of unimportant features. 

Multi-gate Mixture-of-Experts \cite{ma_2018} is a recently proposed shared trunk model,
with a twist: the network contains multiple shared trunks, and each task-specific output
head receives as input a linear combination of the outputs of each shared trunk. The
weights of the linear combination are computed by a separate gating function, which
performs a linear transformation on the network input to compute the linear combination
weights. The gating function can either be shared between all tasks, so that each
task-specific output head receives the same input, or task-specific, so that each output
head receives a different mixture of the shared trunk outputs. This model bears
resemblance to Cross-Stitch networks \cite{misra_2016} (see section \ref{cross_talk}),
but performs a single linear combination of shared components instead of multiple
feature combinations from task-specific layers. This method wasn't empirically evaluated
on computer vision tasks, but is discussed here due to its close relationship with the
other CV architectures \cite{zhang_2014, misra_2016}.

\subsubsection{Cross-Talk} \label{cross_talk}
\begin{figure}[tb]
    \centerline{\includegraphics[scale=0.2]{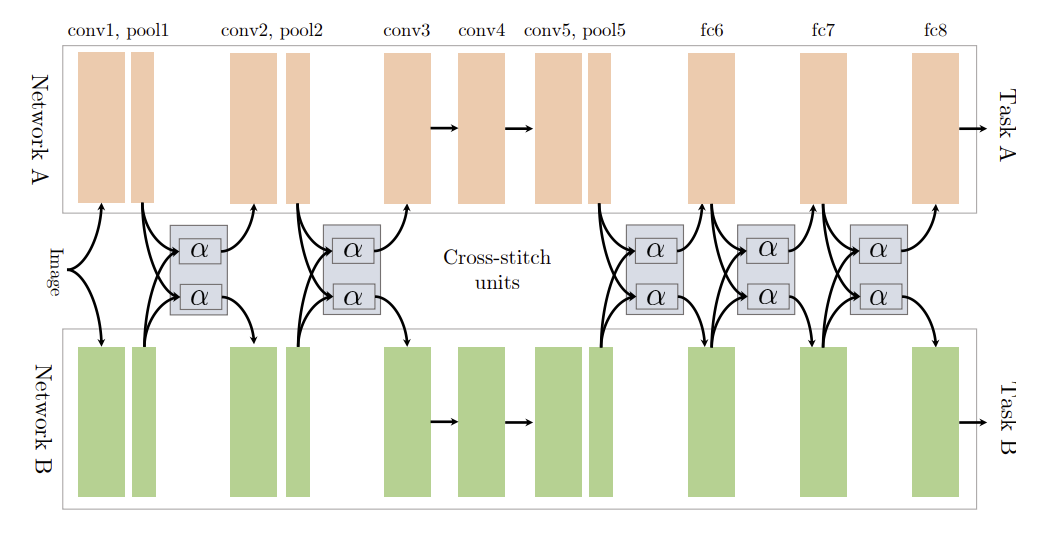}}
    \caption{Cross-Stitch network architecture \cite{misra_2016}. Each task has a
    separate network, but cross-stitch units combine information from parallel layers of
    the different task networks with a linear combination.}
    \label{cross_stitch}
\end{figure}

Not all MTL architectures for computer vision consist of a shared, global feature
extractor with task-specific output branches or modules. \cite{misra_2016, ruder_2019,
gao_2019} take a separate approach. Instead of a single shared extractor, these
architectures have a separate network for each task, with information flow between
parallel layers in the task networks, referred to as \textit{cross-talk}. Figure
\ref{cross_stitch} depicts this idea with the Cross-Stitch network architecture from
\cite{misra_2016}.

\begin{figure}[tb]
    \centerline{\includegraphics[scale=0.165]{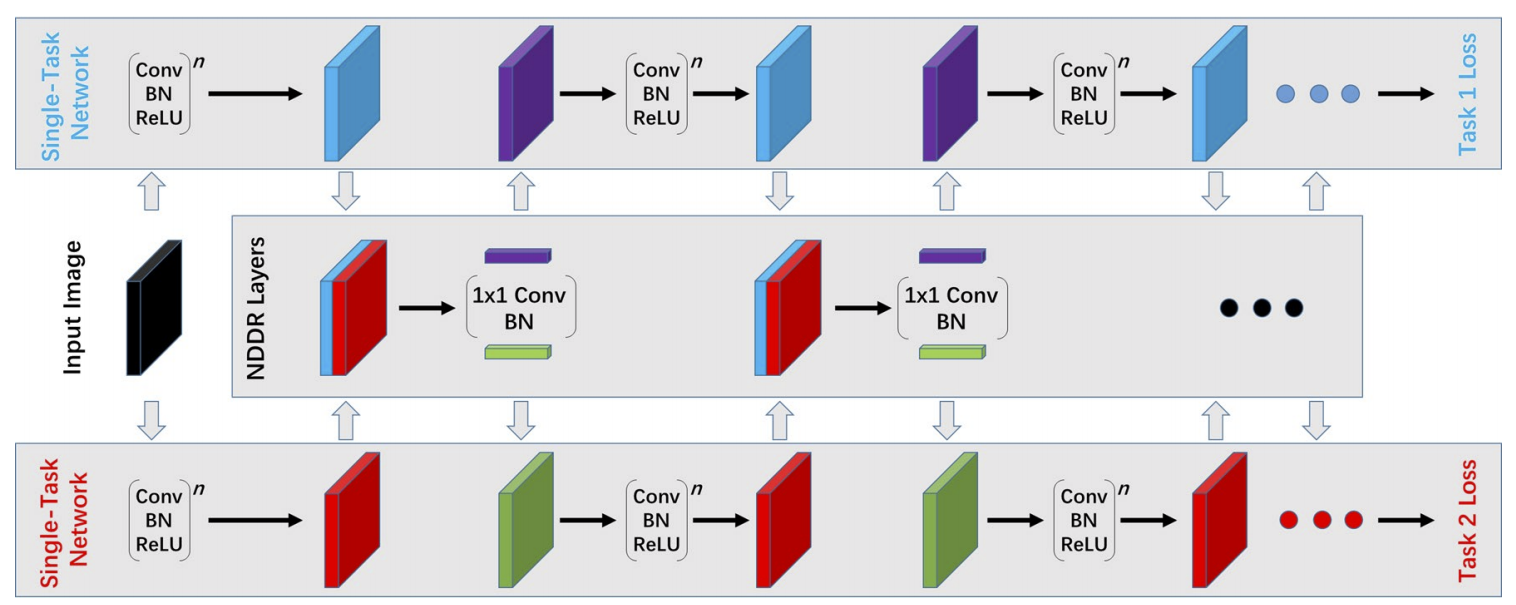}}
    \caption{NDDR-CNN network architecture \cite{gao_2019}. Instead of combining
    information from different task networks with a linear combination of parallel
    features (as in Cross-Stitch networks \cite{misra_2016}), NDDR-CNN uses
    concatenation and a 1x1 convolution to fuse features from separate task networks.}
    \label{nddr_cnn}
\end{figure}

A Cross-Stitch network is composed of individual networks for each task, but the input
to each layer is a linear combination of the outputs of the previous layer from every
task network. The weights of each linear combination are learned and task-specific, so
that each layer can choose which tasks to leverage information from. \cite{ruder_2019}
generalizes this idea with the introduction of the Sluice network. In the Sluice
network, each layer is divided into task-specific and shared subspaces, and the input to
each layer is a linear combination of the task-specific and shared outputs of the
previous layer from each task network. This way, each layer can choose whether to focus
on task specific or shared features from previous layers. The task-specific and shared
subspaces of each layer are also encouraged to be orthogal, by adding an auxiliary term
to the loss function to minimize the squared Frobenius norm of the product of each
task-specific subspace with its corresponding shared subspace. It should be noted that
Sluice networks are presented in a domain-agnostic way, but we discuss them here due to
their relation to Cross-Stitch networks. Finally, \cite{gao_2019} generalizes the
feature fusion operation at parallel layers with Neural Discriminative Dimensionality
Reduction (NDDR-CNN). Instead of using a linear combination to combine features from
parallel layers of the task networks, NDDR-CNN concatenates the outputs from each layer
and pass the result through a 1x1 convolution. The parameters of this convolution are
task specific, as are the linear combination weights in Cross-Stitch networks. A diagram
is shown in figure \ref{nddr_cnn}. Note that this method for feature fusion is a
generalization of Cross-Stitch networks. The 1x1 convolutional parameters can be learned
in such a way to mimic a Cross-Stitch network, but most parameter combinations lead to
feature fusion operations which can't be implemented with a Cross-Stitch network.

\cite{yang_2016b} proposes an architecture which is related to the cross-talk template,
though perhaps only tangentially. In the Sluice network, task-specific and shared
parameter tensors from each layer are simply concatenated to form the layer's
parameters. The architecture of \cite{yang_2016b} also creates an explicit separation
between task-specific and shared parameters, but does so using tensor factorization, a
well-known approach in the classical MTL literature \cite{evgeniou_2004, argyriou_2008,
kumar_2012}. Tensor factorization is used in MTL to represent a multi-task model's
parameter tensor as a product of two smaller tensors, one shared between tasks and one
task-specific, which enforces a different type of division of shared/task-specific
feature spaces than, for example, Sluice networks. \cite{yang_2016b} extends this
approach to the deep learning setting in order to learn the sharing structure at each
layer within a deep network. Unfortunately, there is no empirical comparison of this
tensor factorization approach with the other cross-talk architectures, and there hasn't
been much work extending the tensor factorization approach of \cite{yang_2016b} for deep
MTL.

\subsubsection{Prediction Distillation}
A major tenant and popular justification of MTL is that learned features from one task
may be useful in performing another related task. Prediction distillation techniques are
based on a natural extension of this principle: that the answers to one task may help
learning of another. \cite{vandenhende_2020} provides a great motivating example of this
phenomenon: In an MTL setup for jointly learning depth prediction and semantic
segmentation, discontinuities in the depth map imply likely discontinuities in semantic
segmentation labels, and vice versa. PAD-Net \cite{xu_2018b}, Pattern-Affinitive
Propagation \cite{zhang_2019}, and MTI-Net \cite{vandenhende_2020} each take advantage
of this phenomenon for the multi-task learning of computer vision tasks by making
preliminary predictions for each task, then combining these predictions to produced
final, refined outputs.

\begin{figure}[tb]
    \centerline{\includegraphics[scale=0.2]{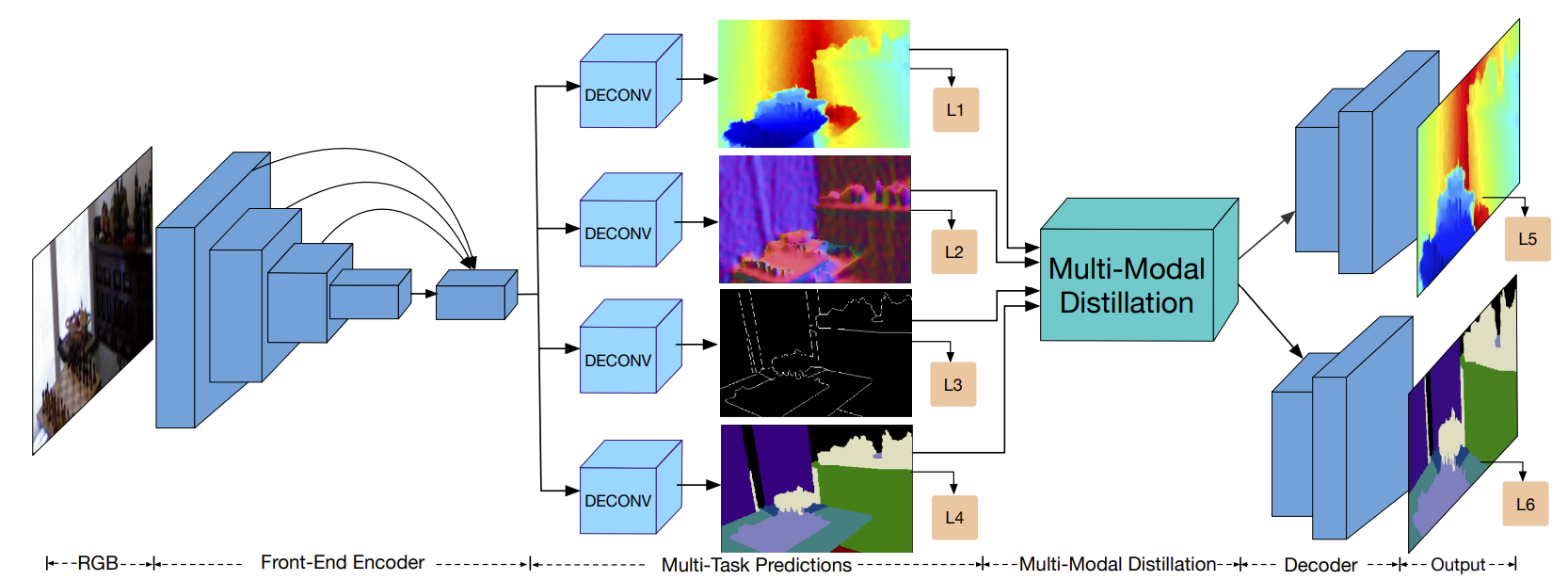}}
    \caption{PAD-Net architecture for prediction distillation \cite{xu_2018b}.
    Preliminary predictions are made for four tasks, then these predictions are
    re-combined and used to compute final, refined predictions for two output tasks.}
    \label{pad_net}
\end{figure}

PAD-Net \cite{xu_2018b} is the earliest of these works, introducing an architecture to
combine preliminary predictions for depth prediction, scene parsing, surface normal
estimation, and contour prediction to produce refined predictions for depth prediction
and scene parsing, as pictured in figure \ref{pad_net}. The preliminary predictions are
recombined using one of three novel variations of a multi-modal distillation module,
either using naive feature concatenation, message passing, or attention-guided message
passing. Pattern-Affinitive Propagation (PAP) \cite{zhang_2019} expands on this
architecture by introducing an affinity learning layer which learns to represent
pair-wise relationships of tasks and combines features from various tasks according to
these relationships. PAP also does away with the extra auxiliary tasks of PAD-Net and
instead produces both preliminary and final predictions for depth prediction, surface
normal estimation, and semantic segmentation. Both of these methods, at the times of
their publication, reached state of the art performance on at least one task from the
NYU-v2 dataset \cite{silberman_2012}.

This style of architecture is further extended by the recently proposed MTI-Net
\cite{vandenhende_2020}, which models task interactions at multiple scales of the
receptive field. Specifically, the architecture consists of a backbone that extracts
multi-scale features, and features from each scale are used to make preliminary task
predictions. The initial predictions from the 1/32 scale are combined with 1/16 scale
features to form the input for predictions at the 1/16 scale, then the predictions from
the 1/16 scale are used as input to make predictions at the 1/8 scale, etc. After
predictions are made from each scale, the predictions are distilled between tasks and
aggregated across scales to make the final refined task predictions. The motivation
behind this multi-scale interaction network comes from the fact that one task's features
or ground-truth outputs may only be informative for learning another task at some (but
not all) scales. The authors consider an example of adjacent cars: at the local level,
when only considering small image patches, the depth discontinuity between cars suggests
that there should be a change in the semantic labels across this discontinuity. At the
global level, however, one can see that the objects surrounding the depth discontinuity
have the same semantic label, which contradicts the supposed task interaction at the
local level. This scenario suggests that multi-scale information should be considered
when distilling predictions across tasks, and indeed this model does show a larger
improvement over single-task counterparts than high-performing baselines.

\subsubsection{Task Routing}
Despite their success, shared trunk and cross-talk architectures are somewhat rigid in
their parameter sharing scheme. \cite{strezoski_2019a} presents an architecture which is
more flexible, allowing for fine-grained parameter sharing between tasks that occurs at
the feature level instead of the layer level. The novel component of this architecture
is the Task Routing Layer which applies a task-specific binary mask to the output of a
convolutional layer to which it is applied, zeroing out a subset of the computed
features and effectively assigning a subnetwork to each task which overlaps with that of
other tasks. The binary masks are not learned, instead they are randomly initialized at
the beginning of training and fixed from that point on. Although this random
initialization doesn't allow for the possibility of a principled parameter sharing
scheme between tasks, the user still has control over the degree of sharing between
tasks through the use of a hyperparameter $\sigma$, known as the sharing ratio. $\sigma$
takes values between 0 and 1, specifying the proportion of units in each layer which are
task-specific, and the random initialization of the binary masks in each layer are
executed in a way to fit this constraint. The proposed architecture only requires a
small increase in the number of parameters as the number of tasks increases, and
experiments demonstrate superior performance over MTL baselines such as the Cross-Stitch
network. Impressively, the Task Routing Layer allows for the network to scale up to
handle up to 312 tasks simultaneously while maintaining decent performance. The Task
Routing Layer is strongly related to the learned architectures Piggyback
\cite{mallya_2018} and Sparse Sharing Architectures \cite{sun_2019b} (discussed in
section \ref{fine_grained_sharing}), though in these works the binary masks which assign
a set of units to each task are learned.

\subsubsection{Single Tasking} \label{single_tasking}
Nearly every multi-task architecture for computer vision produces output for multiple
tasks from the same given input, and each one we have discussed so far satisfies this
condition. \cite{maninis_2019} is, to our knowledge, the only such method which handles
a single task at once, but can be used for multiple tasks with multiple forward passes.
The authors argue that, since the network only performs inference for a single task at a
time, the network is better able to leverage task-specific information and disregard
information useful for other tasks. This focusing is accomplished through the use of two
different attention mechanisms: task-specific data-dependent modulation
\cite{perez_2018} and task-specific Residual Adapter blocks \cite{rebuffi_2018}. The
network is also trained with an adversarial loss \cite{liu_2017} to encourage the
gradients from each task to be indistinguishable. The idea of using an adversarial setup
to encourage similar gradient directions between tasks has also been explored outside of
the realm of computer vision, and is discussed further in section \ref{adv_grad_mod}.

\subsection{Architectures for Natural Language Processing} \label{nlp_architectures}
Natural language processing naturally lends itself well to MTL, due to the abundance of
related questions one can ask about a given piece of text and the task-agnostic
representations which are so often used in modern NLP techniques. The development in
neural architectures for NLP has gone through phases in recent years, with traditional
feed-forward architectures evolving into recurrent models, and recurrent models being
succeeded by attention based architectures. These phases are reflected in the
application of these NLP architectures for MTL.

It should also be noted that many NLP techniques could be considered as multi-task in
that they construct general representations which are task-agnostic (such as word
embeddings), and under this interpretation a discussion of multi-task NLP would include
a large number of methods which are better known as general NLP techniques. Here, for
the sake of practicality, we restrict our discussion to mostly include techniques which
explicitly learn multiple tasks simultaneously for the end goal of performing these
tasks simultaneously.

\subsubsection{Traditional Feed-Forward} \label{feed_forward}
\begin{figure}[tb]
    \centerline{\includegraphics[scale=0.2]{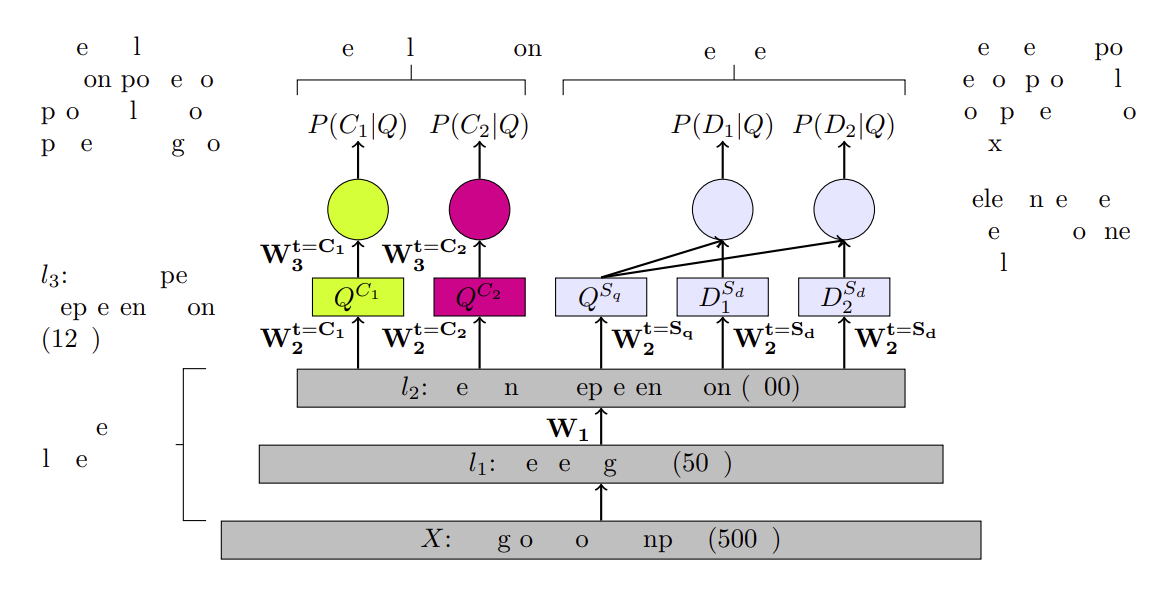}}
    \caption{Network architecture of \cite{liu_2015a}. The input is converted to a
    bag-of-words representation and hashed into letter 3-grams, followed by a shared
    linear transformation and nonlinear activation function. This shared representation
    is passed to task-specific outhead heads to compute final outputs for each task.}
    \label{liu_2015a}
\end{figure}

\cite{collobert_2008, collobert_2011, liu_2015a} all use traditional feed-forward
(non-attention based) architectures for multi-task NLP. Many of these architectures have
a structural resemblance to the early shared architectures of computer vision: a shared,
global feature extractor followed by task-specific output branches. In this case,
however, the features are word representations. \cite{collobert_2008} uses a shared
lookup table layer to learn word representations, where the parameters of each word
vector are directly learned through gradient descent. The remainder of the architecture
is task-specific, and comprised of convolutions, max over time, fully connected layers,
and a softmax output. The seminal work \cite{collobert_2011} is motivated by the general
principles of MTL: representations which are shared across tasks generalize better, and
sharing can improve performance on all tasks. Their architecture is similar to that of
\cite{collobert_2008}, with lookup tables followed by sequences of convolutions and
linear transformations. The main architectural difference is that the first hidden layer
(whether it be linear or convolutional) following the lookup tables is shared between
tasks. Following this trend, the architecture of \cite{liu_2015a} has a similar degree
of sharing, and is pictured in figure \ref{liu_2015a}. In this case word vectors aren't
learned directly. Instead, the input sentence or document is converted into a
bag-of-words representation, and hashed into letter 3-grams. These features are then fed
into a shared linear projection followed by a tanh activation function, and then fed to
task specific output branches.

\subsubsection{Recurrence} \label{recurrence}
The introduction of modern recurrent neural networks for NLP yielded a new family of
models for multi-task NLP, with novel recurrent architectures introduced in
\cite{luong_2015, liu_2016a, liu_2016b, dong_2015}. Sequence to sequence learning
\cite{sutskever_2014} was adapted for multi-task learning in \cite{luong_2015}. In this
work, the authors explore three variants of parameter sharing schemes for multi-task
seq2seq models, which they name one-to-many, many-to-one, and many-to-many. In
one-to-many, the encoder is shared through all tasks, and the decoder is task-specific.
This is useful to handle sets of tasks which require differently formatted output, such
as translating a piece of text into multiple target languages. In many-to-one, the
encoder is task-specific, while the decoder is shared. This is an inversion of the usual
parameter sharing scheme in which earlier layers are shared and feed into task-specific
branches. The many-to-one variant is applicable when the set of tasks require output in
the same format, such as in image captioning and machine translation into the same
target language. Lastly, the authors explore the many-to-many variant, in which there
are multiple shared or task-specific encoders and decoders. They use this variant, for
example, to jointly train an english to german and a german to english translation
system, with both an english and german encoder and decoder. The english encoder also
feeds into the english decoder to perform an autoencoder reconstruction task, as does
the german encoder. A similar sequence to sequence architecture for machine translation
is proposed in \cite{dong_2015} with a focus on training a multi-task network to
translate one source language into multiple target languages.

\begin{figure}[tb]
    \centerline{\includegraphics[scale=0.3]{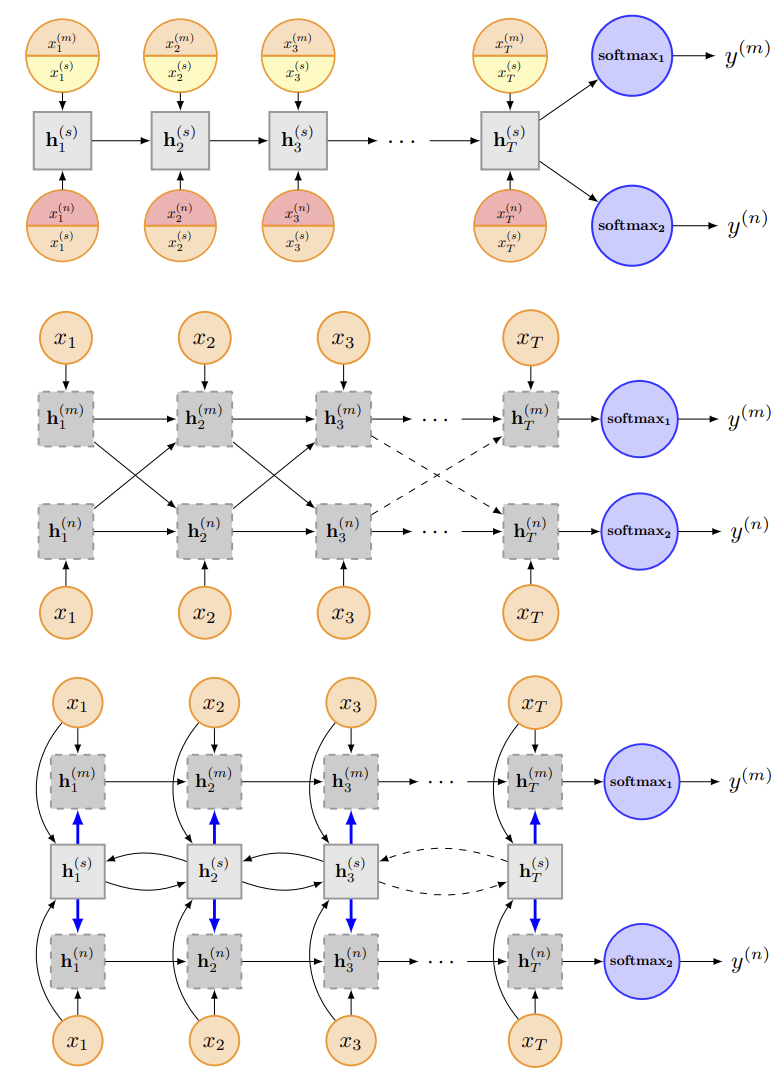}}
    \caption{From top to bottom: Uniform-Layer, Coupled-Layer, and Shared-Layer
        architectures of \cite{liu_2016a}. Each architecture presents a novel partition
        of recurrent architecture components into shared and task-specific modules.}
    \label{liu_2016a}
\end{figure}

\cite{liu_2016a} also explores several variants of recurrent multi-task architectures,
though in the text classification regime instead of sequence to sequence learning. These
parameter sharing schemes are generally more fine-grained than those described in
\cite{luong_2015}, with a focus on different methods to allow information flow betwen
tasks. The authors explore three parameter sharing schemes: the Uniform-Layer,
Coupled-Layer, and Shared-Layer architectures, which are shown in figure
\ref{liu_2016a}. In the Uniform-Layer architecture, each task has its own embedding
layer, and all tasks share an embedding layer and an LSTM layer. Let $i$ be a task
index, $t$ be the recurrent timestep, and $x_t$ be the $t$-th word in an input sentence.
Then the input to the shared LSTM layer for task $i$ on timestep $t$ is the
concatenation of the $i$-th task specific embedding of $x_t$ with the shared embedding
of $x_t$. For the Coupled-Layer model, each task has its own separate LSTM layer, but
each task can read from the LSTM layers of the other tasks. More specifically, the
memory content of the LSTM for a given task at timestep $t$ is modified to include a
weighted sum of the hidden states of the LSTM layers of each task at timestep $t-1$,
while preserving all other components of the LSTM. Finally, the Shared-Layer
architecture allocates a separate LSTM layer for each task, as well as a shared
bi-directional LSTM layer that feeds into the task-specific LSTMs.

In addition to these recurrent architectures, \cite{liu_2016b} augments the LSTM
architecture with a memory mechanism. To form a shared architecture, each task has its
own LSTM parameters, but the memory is shared among all tasks. The memory mechanism is
inspired by the memory enhanced LSTM (ME-LSTM) \cite{sukhbaatar_2015}. The novel
contribution of \cite{liu_2016b} is in a fusion mechanism that allows the memory to be
read from and written to jointly by all tasks. With this addition, the hidden state of
each task's LSTM is computed from a gated sum of the LSTM's internal memory and the
information held in the shared external memory. The authors also introduce a variant in
which each task has its own private external memory, and the shared global external
memory is read/written by each task-specific memory module.

\subsubsection{Cascaded Information} \label{cascade}
\begin{figure}[tb]
    \centerline{\includegraphics[scale=0.25]{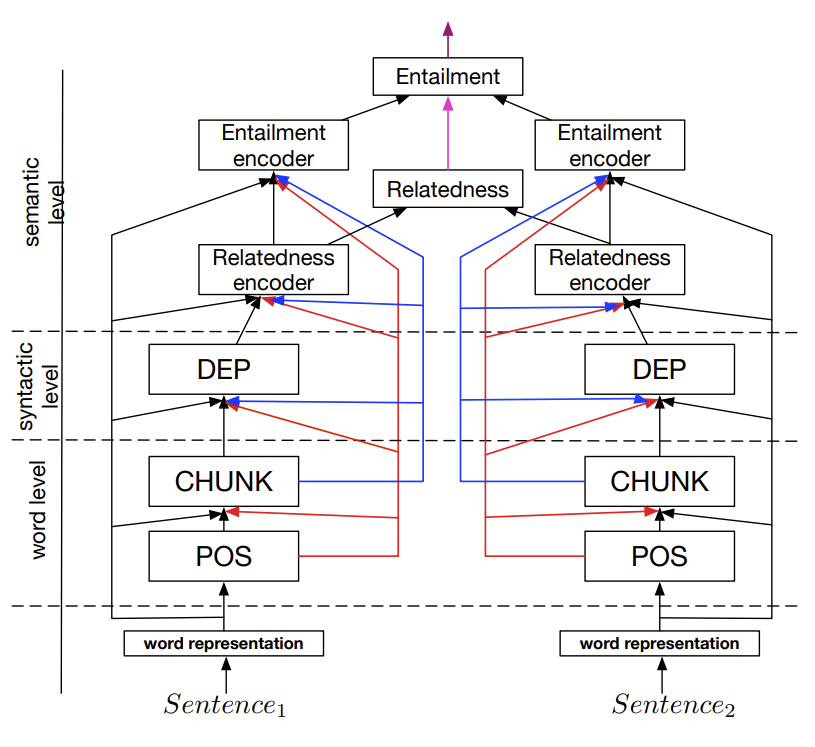}}
    \caption{Various task supervision in various layers from \cite{hashimoto_2016}.
    Lower level tasks are supervised at earlier layers.}
    \label{hashimoto_2016}
\end{figure}

In all of the NLP architectures we have discussed so far, the sub-architectures
corresponding to each task have been symmetric. In particular, the output branch of each
task occurs at the maximum network depth for each task, meaning that supervision for the
task-specific features of each task occurs at the same depth. Several works
\cite{sogaard_2016, hashimoto_2016, sanh_2019} propose supervising ``lower-level" tasks
at earlier layers so that the features learned for these tasks may be used by
higher-level tasks. By doing this we form an explicit task hierarchy, and provide a
direct way for information from one task to aid in the solution of another. We refer to
this template for iterated inference and feature combination as \textit{cascaded
information}, with an example pictured in figure \ref{hashimoto_2016}.

\cite{sogaard_2016} forms this hierarchy by choosing POS tagging as a low-level task to
inform syntactic chunking and CCG supertagging. Their network architecture is made of a
series of bi-directional RNN layers, and for each task $i$ there is an associated layer
$\ell_i$ from which the task-specific classifier for task $i$ stems. In this case, the
associated layer for POS tagging occurs earlier in the network than the associated
layers of syntactic chunking and CCG supertagging, so that the learned POS features can
inform the tasks of syntactic chunking and CCG supertagging. Not long after the
publication of \cite{sogaard_2016}, \cite{hashimoto_2016} achieved a mix of SOTA and
SOTA-competitive results on several language tasks by constructing a similarly
supervised architecture with 5 tasks: POS tagging, chunking, dependency parsing,
semantic relatedness, and textual entailment. The authors also replace the
bi-directional RNN units of \cite{sogaard_2016} with bi-directional LSTM units. Figure
\ref{hashimoto_2016} shows their architecture.

Besides the increase in the number of tasks, this method also introduces a
regularization term to avoid training interference between the tasks. Each time a task's
dataset is sampled for training, the squared Euclidean distance between the pre-update
parameters and the current model parameters is added to the loss function. This
encourages the network parameters not to stray too far from the parameter configuration
which was learned by training on a different task on the previous epoch.

Following these two works, \cite{sanh_2019} introduces a similarly inspired model for a
different set of tasks, achieving SOTA results for Named Entity Recognition, Entity
Mention Detection and Relation Extraction. In order from lowest to highest, the task
hierarchy in this work is NER, EMD, and coreference resolution/relation extraction
(equally ranked as highest level).

\subsubsection{Adversarial Feature Separation} \label{adversarial}
In a novel application of adversarial methods, \cite{liu_2017} introduces an adversarial
learning framework for multi-task learning in order to distill learned features into
task-specific and task-agnostic subspaces. Their architecture is comprised of a single
shared LSTM layer and one task-specific LSTM layer per task. Once the input sentence
from a task is passed through the shared LSTM layer and the task-specific LSTM layer,
the two outputs are concatenated and used as the final features to perform inference on.
However, the features produced by the shared LSTM layer are also fed into the task
discriminator. The task discriminator is a linear transformation followed by a softmax
layer that is trained to predict which task the original input sentence came from. The
shared LSTM layer is then trained to jointly minimize the task loss along with the
discriminator loss, so that the features produced by the shared LSTM do not contain any
task-specific information. In addition, the shared features and the task specific
features are encouraged to encode separate information with the use of an orthogonality
penalty (similar to \cite{ruder_2019}) on the resulting features. More specifically, the
orthogonality loss is defined as the squared Frobenius norm of the product of the
task-specific features and the shared features. This loss is added to the overall
training objective, in order to encourage the task-specific and the shared features to
be orthogonal. These two auxiliary losses enforce the separation of task-specific and
task-agnostic information in the shared network.

\subsubsection{BERT for MTL} \label{bert}
Despite the popularity of the Bidirectional Encoder Representations from Transformers
(BERT) \cite{devlin_2018}, there have been surprisingly little applications of the text
encoding method for MTL. \cite{liu_2019b} extends the work of \cite{liu_2015a} by adding
shared BERT embedding layers into the architecture. The network architecture overall is
quite similar to \cite{liu_2019b}, the only difference being the addition of BERT
contextual embedding layers following the input embedding vectors in figure
\ref{liu_2015a}. This new MTL architecture, named MT-DNN, achieved SOTA performance on
eight out of nine GLUE tasks \cite{wang_2018} at the time of its publication.

\subsection{Architectures for Reinforcement Learning} \label{rl_architectures}
In recent years, many of the advances in reinforcement learning have focused on
optimization and training methods \cite{schulman_2017, haarnoja_2018, akkaya_2019}.
Since many RL problems don't necessarily involve complex perception, such as working
with words or pixels, the architectural demand isn't as high for many RL problems.
Because of this, many deep networks for RL are simple fully-connected, convolutional, or
recurrent architectures. However, in the multi-task case, there are several instances of
interesting works which leverage information between tasks to create improved
architectures for RL.

\subsubsection{Joint Task Training} \label{joint_training}
\begin{figure}[tb]
    \centerline{\includegraphics[scale=0.25]{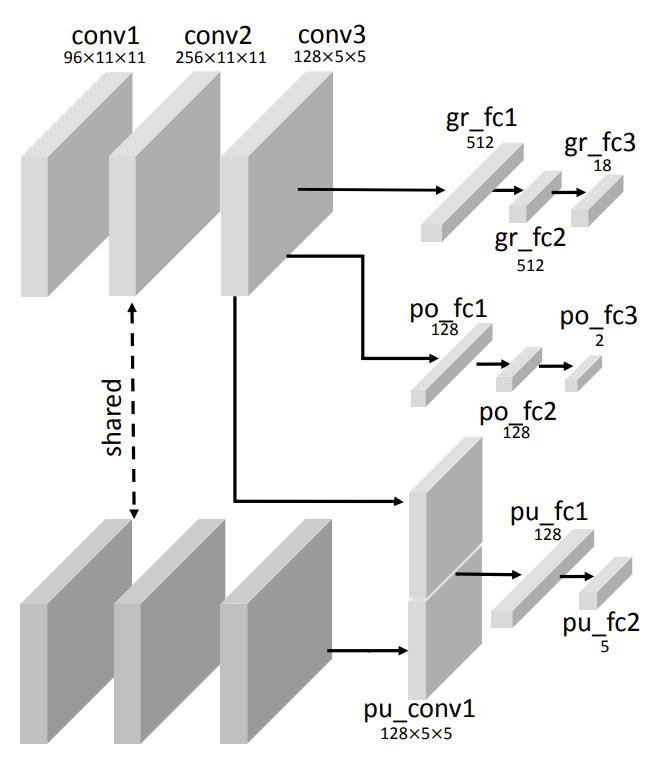}}
    \caption{Shared architecture for robotic grasping, pushing, and poking
        \cite{pinto_2017}.}
    \label{pinto_2017}
\end{figure}

Several works in RL have found that task performance can be improved by simply training
for multiple tasks jointly, with or without parameter sharing. \cite{pinto_2017} uses a
shared trunk architecture (shown in figure \ref{pinto_2017} to jointly learn robotic
grasping, pushing, and poking from pixels. The shared feature extractor consists of
three convolutional layers, and these shared features are fed to three task-specific
output branches. The grasping and poking output branches are made of three
fully-connected layers each, and the pushing branch has one convolutional layer,
followed by two fully-connected layers. This shared network is trained with a supervised
loss which is an average of cross-entropy and squared Euclidean losses, one for each
task. The network actions are parameterized in such a way to allow for supervised
training. The authors find that this shared network trained with 2500 examples of both
pushing and grasping outperforms a task-specific grasping network trained with 5000
examples. \cite{zeng_2018} also finds advantages by jointly training with robotic
pushing and grasping, though their architecture does not employ any parameter sharing.
The network is comprised of two separate fully convolutional Q-networks, one for pushing
and one for grasping. The two networks are, however, given a joint training signal. The
reward for a timestep $t$ is defined as follows: if a grasping action is chosen at
timestep $t$ and the grasp is successful, the reward is 1. If a pushing action is chosen
at timestep $t$ and the action causes a sufficiently large change in the environment,
then the reward is 0.5. From this reward, there is no explicit encouragement of one task
to aid another. But when both networks are jointly trained to maximize the same reward,
the pushing network learns to push in a way that influences the environment to maximize
the grasping reward. This joint training setup was shown to be much more sample
efficient than baselines, making training on a physical robot feasible with only a few
hours of training.

\subsubsection{Modular Policies} \label{modular_policies}
There have been many similarities between the various parameter sharing schemes that we
have discussed so far, but modular networks are present a novel family of parameter
sharing methods which are totally different from the shared trunk or cross-talk
architectures from sections \ref{shared_trunk} and \ref{cross_talk}. In modular learning
setups, each task's network architecture is composed of a combination of smaller
sub-networks, and these smaller sub-networks are combined in different ways for
different tasks. Just as MTL is motivated by generality through shared representations,
modular learning offers generality of computation through shared neural building blocks.
The goal of these setups is to learn building blocks which are general enough to be
useful as a part of the network architecture for multiple tasks. We discuss several
other learned modular architectures in sections \ref{learned_architectures} and
\ref{conditional_architectures}, but here we only discuss those modular methods for
which the parameters of the building blocks are learned and the configuration of
building blocks for each task remains fixed. Discussion of modular methods with learned
building block combinations can be found in the aforementioned sections.

\begin{figure}[tb]
    \centerline{\includegraphics[scale=0.35]{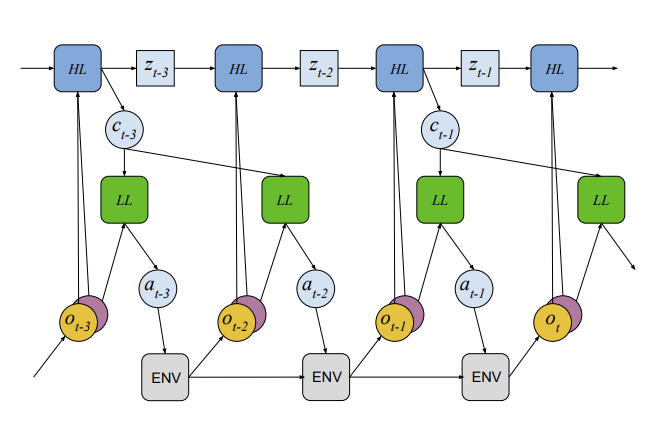}}
    \caption{Shared modular architecture for locomotion with multiple robots
        \cite{heess_2016}. Note that the high-level module updates the modulated input to
        the low-level module at a different frequency than it itself receives input from the
        environment.}
    \label{heess_2016}
\end{figure}

\begin{figure}[tb]
    \centerline{\includegraphics[scale=0.15]{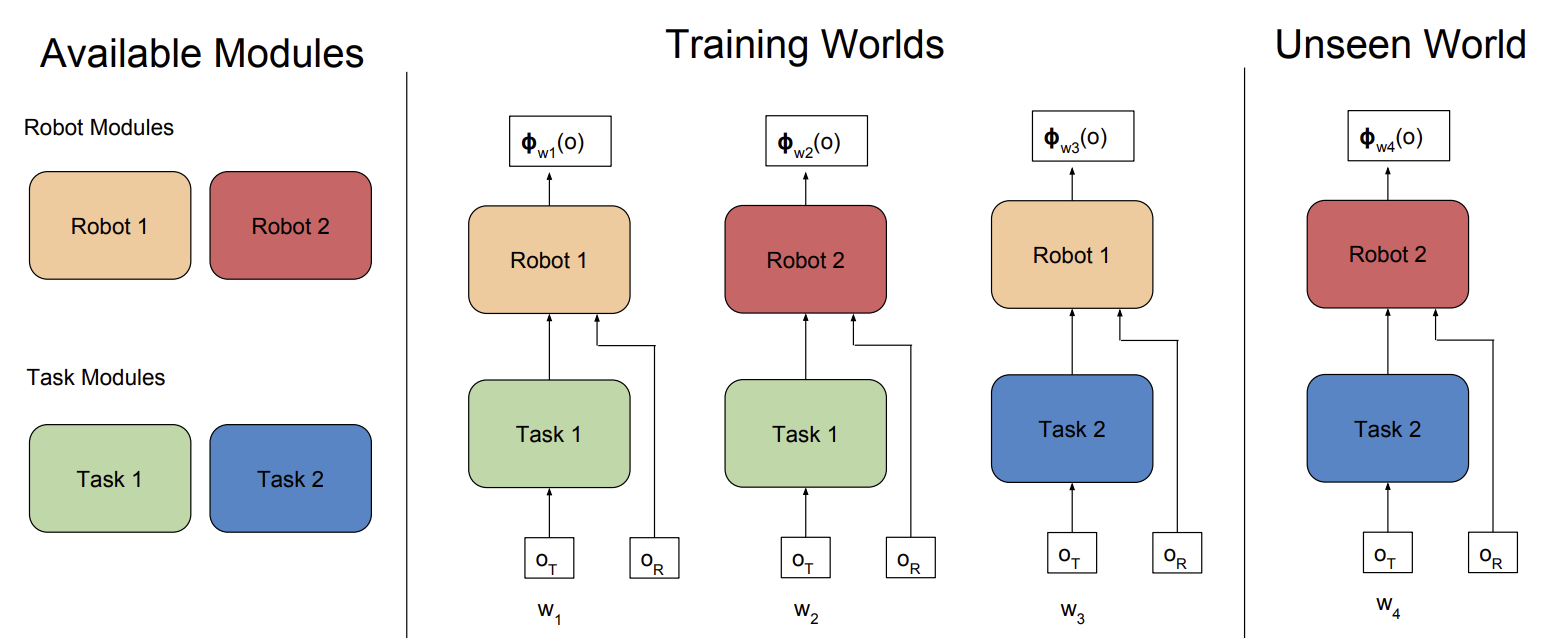}}
    \caption{Shared modular architecture for multi-task and multi-robot transfer
    \cite{devin_2017}. Each network is made of two modules, one robot module and one
    task module. Each robot module can be combined with a task module to form a network
    to perform each (task, robot) pair.}
    \label{devin_2017}
\end{figure}

Within weeks of each other, \cite{heess_2016} and \cite{devin_2017} both introduced
modular neural network policies for multi-task learning across various robots. The
architectures of each of these works are depicted in figures \ref{heess_2016} and
\ref{devin_2017}, respectively.

The task architecture of \cite{heess_2016} is made of two modules, a low-level
``spinal" network and a high-level ``cortical" network. The spinal network has access to
proprioceptive information like muscle tension, and it chooses motor actions, while the
cortical network has access to all observations and modulates inputs to the spinal
network. It is important to note that the proprioceptive information given to the spinal
network is always task-independent, so that the spinal network must learn
task-independent representations. In their experiments, the low-level/spinal network is
feed-forward, while the high-level/cortical network is recurrent. The combination of the
division of labor between the two modules and the information hiding from the spinal
network allows for a pre-trained spinal network to be deployed with a new cortical
network to quickly solve a new task with the same robot body. The usage of the
pre-trained spinal network allows for effective exploration in the robot body, despite
the new task.

The architecture of \cite{devin_2017} is similarly inspired, but employs parameter
sharing for the network controllers between different robots as well as between tasks.
Each task and robot has its own network module. The network for each task/robot pair is
composed of the corresponding task-specific module, followed by the corresponding
robot-specific module, as shown in figure \ref{devin_2017} Because each module is shared
between tasks and robots, it is constrained to learn information which is general across
its domains. The authors also show that the learned modules can be paired in
combinations unseen during training, to instantiate a policy with zero-shot
generalization capabilities. This method also gives partial information to the task
module; each observation is decomposed into a task-specific portion and a robot-specific
portion. The task-specific module only receives the task-specific observation as input,
and the robot-specific module receives the robot-specific observation as well as the
output of the task-specific module.

Both of these architectures exhibit an interesting strategy for learning general
representations across tasks: \textit{information hiding}. We have so far discussed
parameter sharing, adversarial methods, and orthogonality constraints as regularization
strategies for multi-task methods. But the division of labor brought forth by the
modularity in these two architectures allows for information to be restricted to certain
modules in the network, forcing the modules missing this information to learn
representations which are invariant to the omitted information. In this case, we obtain
modules which are invariant to the task at hand.

RL with Policy Sketches \cite{andreas_2017} is another template for policy modularity
which was proposed soon after \cite{heess_2016} and \cite{devin_2017}, in which the
policy for a task is composed of several subpolicies, and each subpolicy is a neural
network whose parameters are shared between tasks. The composition of subpolicies for
each task is defined by a human-provided ``policy sketch", which roughly outlines the
steps to complete a task. For example, in the minecraft-inspired environment used for
evaluation in the paper, the tasks ``Make Planks" and ``Make Sticks" may have the policy
sketches (get wood, use workbench) and (get wood, use toolshed), respectively. In this
case, the policies for these tasks would use the subpolicies $\pi_{\text{wood}}$,
$\pi_{\text{bench}}$, and $\pi_{\text{shed}}$, with the compositions
$\pi_{\text{planks}} = (\pi_{\text{wood}}, \pi_{\text{bench}})$ and $\pi_{\text{sticks}}
= (\pi_{\text{wood}}, \pi_{\text{shed}})$. The weak supervision provided by the policy
sketches defines a sharing structure of subpolicies between tasks, which was shown to be
more beneficial to learning than unsupervised option discovery. This process is similar
to how the syntactical strcture of a question defines the composition of subpolicies in
Neural Module Networks \cite{andreas_2016} (discussed in section
\ref{conditional_architectures}), though in that example the composed architecture is
dependent on each individual given question, while the composition of subpolicies
remains fixed for each task with Policy Sketches. It is important to note that module
composition takes two different forms with Policy Sketches and the architectures of
\cite{heess_2016} and \cite{devin_2017}: subpolicies in Policy Sketches behave as in
hierarchical reinforcement learning \cite{kulkarni_2016}, where a subpolicy is chosen to
act as the policy until some termination condition is met, as opposed to composition in
the sense of function composition as in \cite{heess_2016} and \cite{devin_2017}.

\subsubsection{Multiple Auxiliary Tasks} \label{auxiliary_tasks}
\cite{jaderberg_2016} introduces several unsupervised auxiliary tasks to be learned in
conjuction with a main task, as an additional form of supervision. These auxiliary tasks
encourage general representations in the usual sense for MTL, but they also help to
decrease the sparsity of rewards in the original task.  The architecture is a CNN-LSTM
actor-critic with a shared trunk, and output branches for each auxiliary task that
requires its own output. The auxiliary tasks themselves are called pixel control,
feature control, and reward prediction. Pixel control shares parameters from the agent
CNN and LSTM, and branches off into a task-specific branch that chooses its own actions.
The actions are rewarded for causing maximal change in the pixel intesity of the pixels
observed as a result of the chosen action. Feature control does not require an output,
and instead the agent is rewarded for activating the hidden units of a given hidden
layer of the agent network. Lastly, reward prediction uses the agent's CNN to map three
recent frames to a prediction of the reward on the next step. These auxiliary tasks do
not require any supervision that isn't provided by the environment dynamics and are
general enough to apply to many different problem settings. Training an agent with these
simple auxiliary tasks led to SOTA performance on the Arcade Learning Environment
\cite{bellemare_2013}.

\subsection{Multi-Modal Architectures} \label{multi_architectures}
In sections \ref{cv_architectures}, \ref{nlp_architectures}, and \ref{rl_architectures},
we discussed the multi-task architectures which were specifically designed to handle
data in one fixed domain. Here, we describe architectures to handle multiple tasks using
data from multiple domains, which is usually some combination of visual and linguistic
data. Multi-modal learning is an interesting extension of many of the motivating
principles behind multi-task learning: sharing representations across domains decreases
overfitting and increases data efficiency. In the multi-task single modality case, the
representations are shared across tasks but within in a single modality. However, in the
multi-task multi-modal case, representations are shared across tasks and across modes,
providing another layer of abstraction through which the learned representations must
generalize. This suggests that multi-task multi-modal learning may yield an increase in
the benefits already exhibited by multi-task learning.

\cite{nguyen_2019} introduces an architecture for shared vision and language tasks by
using dense co-attention layers \cite{nguyen_2018}, in which tasks are organized into a
hierarchy and low-level tasks are supervised at earlier layers in the network. Dense
co-attention layers were developed for visual question answering, specifically for the
integration of visual and linguistic information. This setup of task supervision is
similar to the cascaded information architectures discussed in section \ref{cascade}.
However, instead of hand-designing a hierarchy of tasks, this method performs a search
over the layers for each task in order to learn the task hierarchy. The architecture of
\cite{akhtar_2019} handles visual, audio, and text input to classify emotion and
sentiment in a video of a human speaker, using bi-directional GRU layers along with
pairwise attention mechanisms for each pair of modes to learn a shared representation
incorporating all modes of input.

\begin{figure}[tb]
    \centerline{\includegraphics[scale=0.25]{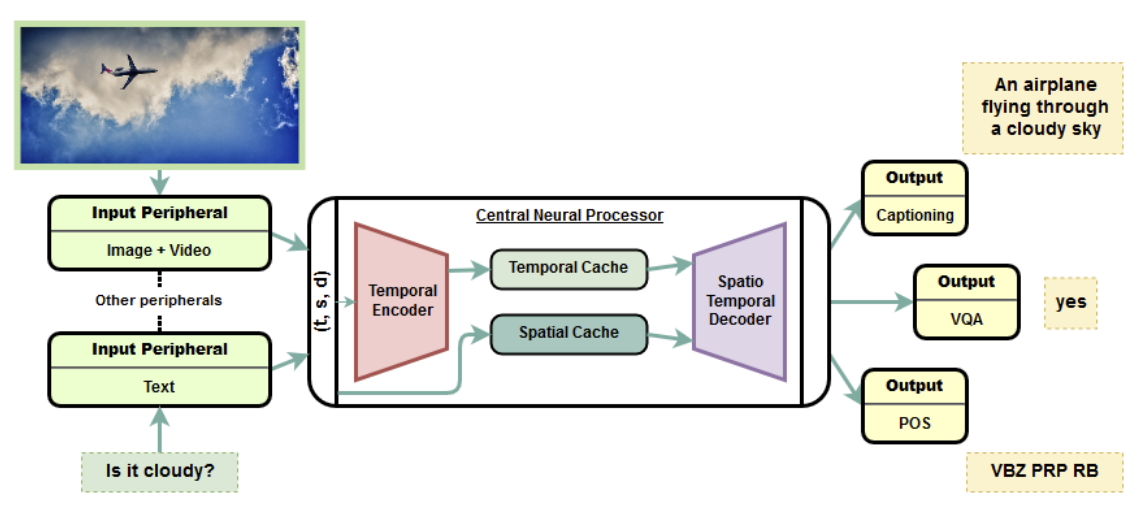}}
    \caption{OmniNet architecture proposed in \cite{pramanik_2019}. Each modality has a
    separate network to handle inputs, and the aggregated outputs are processed by an
    encoder-decoder called the Central Neural Processor. The output of the CNP is then
    passed to several task-specific output heads.}
    \label{omninet}
\end{figure}

Both \cite{nguyen_2019, akhtar_2019} are focused on a set of tasks which all share the
same fixed set of modalities. Instead, \cite{kaiser_2017} and \cite{pramanik_2019} focus
on building a ``universal multi-modal multi-task model", in which a single model can
handle multiple tasks with varying input domains. The architecture introduced in
\cite{kaiser_2017} is comprised of an input encoder, an I/O mixer, and an autoregressive
decoder. Each of these three blocks is made of a mix of convolutions, attention layers,
and sparsely-gated mixture-of-experts layers. The authors also demonstrate that the
large degree of sharing between tasks yields significantly increased performance for
tasks with limited training data. Instead of aggregating mechanisms from various modes
of deep learning, \cite{pramanik_2019} introduces an architecture called OmniNet with a
spatio-temporal cache mechanism to learn dependencies across spatial dimensions of data
as well as the temporal dimension.  A diagram is shown in figure \ref{omninet}. Each
input modality has a corresponding ``peripheral" network, and the outputs of these
networks are aggregated and fed into the Central Neural Processor, whose output is fed
to task-specific output heads. The CNP has an encoder-decoder architecture with a
spatial cache and a temporal cache. OmniNet reached SOTA-competitive performance on POS
tagging, image captioning, visual question answering, and video activity recognition.

Most recently, \cite{lu_2020} introduces a multi-task model that handles 12 different
datasets simultaneously, aptly named 12-in-1. Their model achieves superior performance
over the corresponding single-task models on 11 out of 12 of these tasks, and using
multi-task training as a pre-training step leads to SOTA performance on 7 of these
tasks. The architecture is based on the ViLBERT model \cite{lu_2020}, and is trained
using a mix of methods such as dynamic task scheduling, curriculum learning, and
hyperparameter heuristics.

\subsection{Learned Architectures} \label{learned_architectures}
As we have already seen in the preceding sections, there have been many developments in
the design of shared architectures to emphasize the strengths of multi-task learning
while mitigating the weaknesses. Another approach to architecture design for multi-task
learning is to learn the architecture as well as the weights of the resulting model.
Many of the following methods for learning shared architectures allow for the model to
learn how parameters should be shared between tasks. With a varying parameter sharing
scheme, the model can shift the overlap between tasks in such a way that similar tasks
have a higher degree of sharing than unrelated tasks. This is one potential method for
mitigating negative transfer between tasks: if two tasks exhibit negative transfer, the
model may learn to keep the parameters for those tasks separate. Going further, it may
be the case that two tasks exhibit positive transfer in some parts of the network, and
negative transfer in others. In this case, designing a parameter sharing scheme by hand
to accommodate various task similarites at different parts of the network becomes
infeasible, especially as the number of tasks and the size of the network grows. Learned
parameter sharing offers a way to facilitate adaptive sharing between tasks to a level
of precision that isn't realistic for hand designed shared architectures.

We roughly categorize the methods for learned architectures into four groups:
architecture search, branched sharing, modular sharing, and fine-grained sharing. The
boundaries between these groups aren't concrete, and they are often blurred, but we
believe this is a useful way to broadly characterize the patterns in the recently
developed methods. \textit{Branched sharing} methods are a coarse-grained way to share
parameters between tasks. Once the computation graphs for two tasks differ, they never
rejoin (see figure \ref{lu_2017}). \textit{Modular sharing} represents a more
fine-grained approach, in which a set of neural network modules is shared between tasks,
where the architecture for each task is made by a task-specific combination of some or
all of the modules, as in figure \ref{adashare}. Lastly, the most fine-grained approach
to parameter sharing is what we simply call \textit{fine-grained sharing}, in which
sharing decisions occur at the parameter level instead of the layer level, as shown in
figure \ref{sun_2019b}.

\subsubsection{Architecture Search} \label{architecture_search}
Each of \cite{wong_2017, liang_2018, gao_2020} introduces a method for multi-task
architecture search, but with completely different approaches. \cite{wong_2017}
introduces the Multi-task Neural Model Search (MNMS) controller. This method doesn't
involve a single network which is shared between all tasks. Instead, the MNMS controller
is trained simultaneously on all tasks to generate one individual architecture for each
task. The method is an extension of \cite{zoph_2016}, where an RNN controller
iteratively makes architecture design choices, and is trained with reinforcement
learning to maximize the expected performance of the resulting network. In the
multi-task variant, the RNN also uses task embeddings, which are learned jointly with
the MNMS controller, to condition architectural design choices on the nature of the
task.

On the other hand, \cite{liang_2018} introduces several variations of a multi-task
neural architecture search algorithm that uses evolutionary strategies to learn neural
network modules which can be reordered differently for various tasks. This method is an
extension of the Soft Layer Ordering introduced in \cite{meyerson_2017} (discussed in
section \ref{modular_policies}). Just as in \cite{meyerson_2017}, the method of
\cite{liang_2018} involves learning neural network modules jointly with their ordering
for various tasks. In the architecture search extension, the architecture of the modules
is learned along with their routing for individual tasks. The most sophisticated variant
of this algorithm is called Coevolution of Modules and Task Routing (CMTR), in which the
CoDeepNEAT algorithm \cite{miikkulainen_2019} is used to evolve the architecture of the
shared group modules in an outer loop, and the task specific routings of these modules
are evolved in an inner loop.

Most recently, \cite{gao_2020} proposes MTL-NAS as a method for gradient-based
architecture search in MTL. All architectures in this search space are made of a set of
fixed-architecture single-task backbone networks, one for each task, and the search
process operates over feature fusion operations between different layers of these
single-task networks. The feature fusion operations are parameterized by NDDR (from
NDDR-CNN \cite{gao_2019}, see section \ref{cross_talk}), which is essentially a $1
\times 1$ convolution acting on concatenations of feature maps from different tasks.
This method also introduces a minimum entropy objective on the weights of the fusion
operations, so that the search process converges to a discrete architecture during the
architecture search phase, which diminishes the need for a discretization of a soft
combination of architectures as in other NAS works \cite{liu_2018} and closes the
performance gap between learned soft architectures and the final discretized version.
The final learned architectures were shown to outperform common multi-task baselines on
the NYU-v2 \cite{silberman_2012} and Taskonomy \cite{zamir_2018} datasets.

\subsubsection{Branched Sharing} \label{branched_sharing}
\begin{figure}[tb]
    \centerline{\includegraphics[scale=0.275]{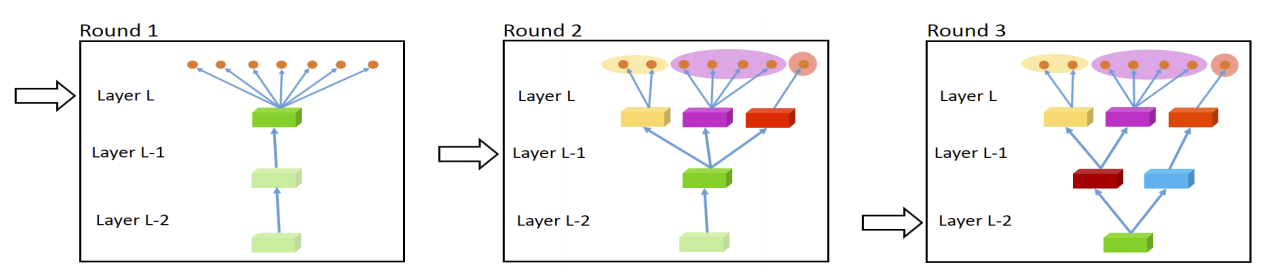}}
    \caption{Learned branching architecture proposed in \cite{lu_2017}. At the beginning
    of training, each task shares all layers of the network. As training goes on, less
    related tasks branch into clusters, so that only highly related tasks share as many
    parameters.}
    \label{lu_2017}
\end{figure}

\cite{lu_2017} is one of the earliest methods for learned parameter sharing in
multi-task deep learning. The idea is to start with a network which is shared between
all tasks up to task-specific output heads, then iteratively decouple parameters between
tasks layer by layer, starting with the layer closest to the output heads, and moving to
the earlier layers. A diagram of this process is shown in figure \ref{lu_2017}. When a
shared layer splits into multiple task specific layers, tasks are clustered based on an
estimate of pairwise task affinity. These task affinities are computed according to the
following principle: two tasks are likely related if the same input data is equally
easy/difficult for the models corresponding to each task.

More recently, \cite{vandenhende_2019} proposes a similar method with a different
criterion for task grouping. Instead of concurrent sample difficulty, this algorithm
uses representation similarity analysis (RSA) \cite{kriegeskorte_2008} as a measure of
task affinity. RSA is built on the principle that similar tasks will rely on similar
features of the input, and will therefore learn similar feature representations. The
other important difference between these methods is that \cite{vandenhende_2019}
computes the branching structure globally instead of greedily by layer. However, the
search over all branching structures is computationally expensive, so the authors resort
to a beam search strategy for computing the branching structure from the representation
similarities across tasks in different parts of the network. This paper includes a
direct comparison of the two methods, and the RSA-based variant is shown to be superior.
RSA is also used in some methods to learn explicit task relationships, which are
discussed in section \ref{relationship}.

\subsubsection{Modular Sharing} \label{modular_sharing}
\begin{figure}[tb]
    \centerline{\includegraphics[scale=0.175]{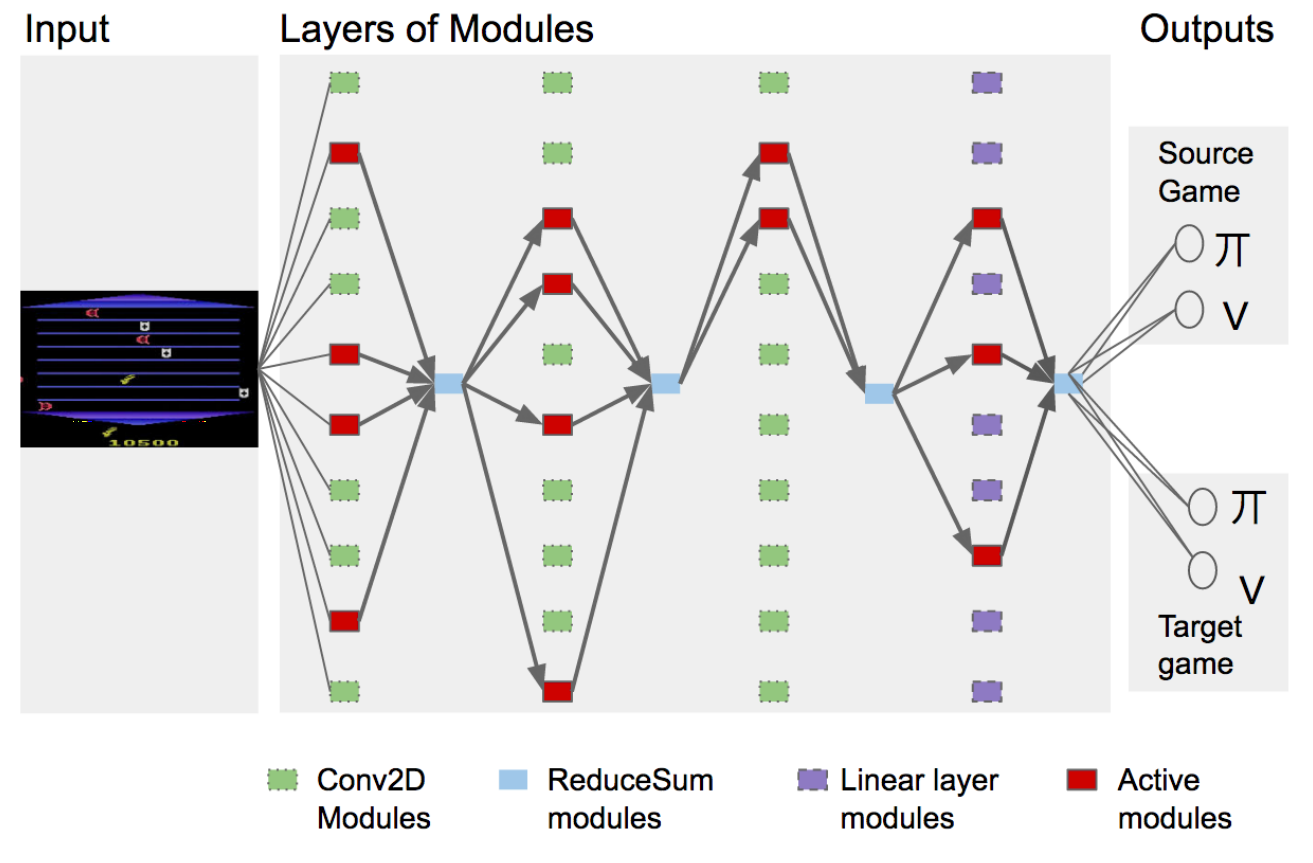}}
    \caption{Example PathNet architecture \cite{fernando_2017}. A large network is
    shared by many tasks, but each task only uses a subnetwork which is evolved through
    a tournament selection genetic algorithm.}
    \label{pathnet}
\end{figure}

The earliest work on modular parameter sharing in multi-task learning that we are aware
of is PathNet \cite{fernando_2017}. A PathNet model is one large neural network which is
used for multiple tasks, though different tasks have different computation pathways
within the larger model. A diagram is shown in figure \ref{pathnet}. The pathway for
each task is learned through a tournament selection genetic algorithm, in which many
different candidate pathways compete and evolve towards an optimal subnetwork of the
larger network. While this idea is mostly general and can be applied to various
settings, such as multi-task learning and meta-learning, the authors deploy this model
for continual learning with two reinforcement learning tasks.  The weights learned
during training on the first task are fixed during training of the second task, during
which new pathways through the network are evolved to complete the task at hand.

\begin{figure}[tb]
    \centerline{\includegraphics[scale=0.2]{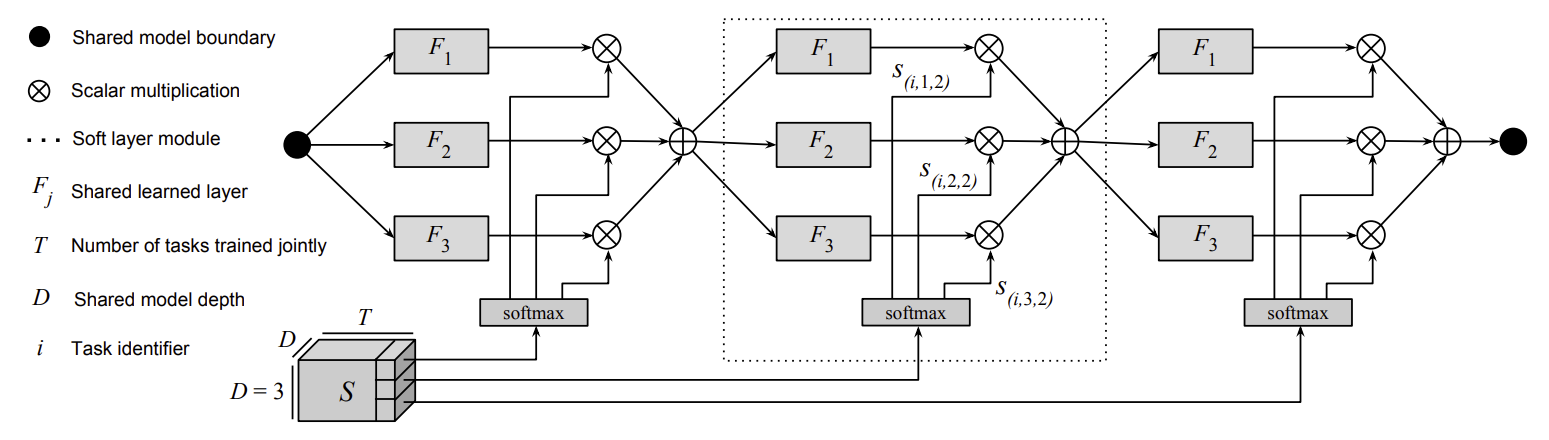}}
    \caption{Soft Layer Ordering with three learned layers \cite{meyerson_2017}. Each
    layer of the network is a linear combination of several network modules, and the
    weights of these combinations are task-specific.}
    \label{soft_layer}
\end{figure}

Soft Layer Ordering \cite{meyerson_2017} and Modular Meta-Learning \cite{alet_2018} are
two concurrent works of modular MTL, similar but with an important difference. Each of
these methods learns a shared set of neural network modules which are combined in
different ways for different tasks, with the hope that a network ``building block" will
learn generally applicable knowledge if it is used in various contexts within the
different task networks. Soft Layer Ordering parameterizes a task network by computing a
convex combination of each module's output at each layer of the network, as shown in
figure \ref{soft_layer}. With this parameterization, each learned module can contribute
to each level of depth in the network. In contrast, Modular Meta-Learning learns a
computation graph over the modules, meaning that each step of the computation is a
discrete composition of a small number of modules, instead of a soft combination of all
of them. The difference in the parameterization of the computation graph between these
methods leads to different optimization strategies, namely, the computation graph in
Soft Layer Ordering architectures can be optimized with gradient descent jointly with
the network weights, since the composition of the modules is a differentiable operation.
In comparison, the computation graph in Modular Meta-Learning is a discrete structure,
so gradient-based optimization methods cannot be used to learn the graph over modules
for each task. Instead, the authors employ simulated annealing, a black box optimization
method, to learn the computation graph. While this two-level optimization incurs
computational cost, the discrete nature of the computation graph affords the ability to
produce an inductive bias in the resulting model, which the soft sharing of layers does
not exhibit. These methods represent two realizations of a broadly generalizable
template that many other methods have employed: Learn individual network pieces, and
learn how to combine them.

The method of \cite{chen_2018} is another strategy in this spirit, and closely resembles
NAS \cite{zoph_2016}. This paper proposes a method that fits the template described
above, but the composition of modules is not directly parameterized and learned such as
in Soft Layer Ordering and Modular Meta-Learning. Instead, this method trains an RNN
controller to choose a layer from a fixed set of layers to iteratively build an
architecture as a sequence of modules, and the module is again trained with
reinforcement learning to maximize the expected performance of the constructed
architecture. This method bears a strong resemblance to the previously discussed
Multi-task Neural Model Search controller \cite{wong_2017}, with the main difference
being that the RNN controller used in \cite{chen_2018} simply chooses between a set of
network modules, while the MNMS controller makes architectural design decisions.

\begin{figure}[tb]
    \centerline{\includegraphics[scale=0.2]{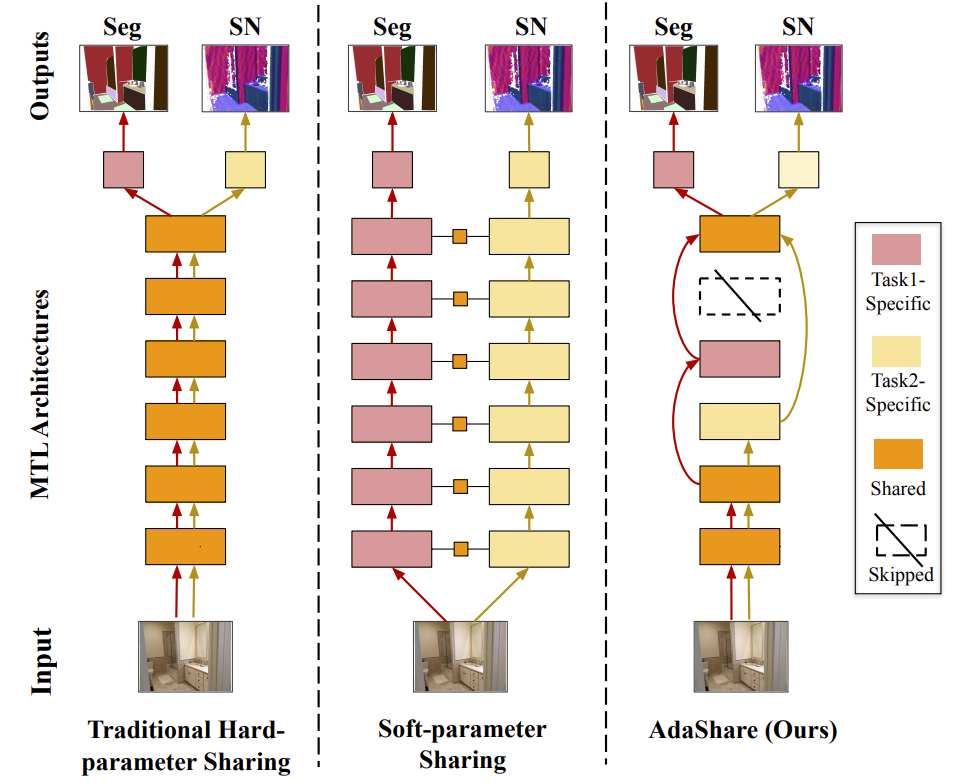}}
    \caption{A learned parameter sharing scheme with AdaShare \cite{sun_2019a}. Each
    layer in the network is either included or ignored by each task, so that each task
    uses a subnetwork which is (likely) overlapping with other tasks.}
    \label{adashare}
\end{figure}

Most recently, AdaShare \cite{sun_2019a} is an algorithm for modular MTL in which each
task architecture is comprised of a sequence of network layers. Each layer in the shared
set is either included or omitted from the network for each task. An example is shown in
figure \ref{adashare}. Along with the weights of each layer, AdaShare learns an $N
\times L$ array of binary values, where $N$ is the number of tasks, $L$ is the total
number of shared layers, and the $(i, \ell)$-th element of the binary array denotes
whether layer $\ell$ is included in the model of the $i$-th task. Since the output of a
task's network is not differentiable with respect to these binary values, the method
adopts Gumbel-Softmax sampling \cite{jang_2016} to optimize these parameters with
gradient descent jointly with the network weights. This strategy makes an interesting
medium between Soft Layer Ordering \cite{meyerson_2017} and Modular Meta-Learning
\cite{alet_2018}, in which each shared module is shared discretely instead of softly,
but the computation graph can still be learned with gradient descent. AdaShare also
employs several regularization terms to encourage sharing in the lower-level modules and
sparsity in the resulting task-specific networks, which are discussed in section
\ref{regularization}.

\subsubsection{Fine-Grained Sharing} \label{fine_grained_sharing}
Fine-grained parameter sharing schemes are the most recently introduced MTL architecture
type, and they allow for more flexible information flow between tasks than sharing at
the layer or multi-layer level. Piggyback \cite{mallya_2018} is a method for adapting a
pre-trained network on a related task by learning to mask out individual weights of the
original network. This allows for the storage of a newly trained model with a storage
cost of only one additional bit per parameter of the original model while preserving the
original network function. Despite the fact that the network output is not
differentiable with respect to these network masks, these network masks are optimized
through gradient descent jointly with the network weights by using a continuous
relaxation of the mask values as a noisy estimate of the binary mask values. This method
of optimizing such mask values is justified in prior work on binarized neural networks
\cite{courbariaux_2015}.

\cite{newell_2019} and \cite{bragman_2019} are concurrent works that each propose a
parameter sharing scheme for multi-task CNNs in which sharing occurs at the filter
level. For each convolutional layer of a multi-task network, the method of
\cite{newell_2019} learns a binary valued $N \times C$ array $M$, where $N$ is again the
number of tasks and $C$ is the number of feature channels in a given layer of the
network. The $(i, c)$-th element of $M$ denotes whether the model of the $i$-th task
should include the $c$-th feature map in the considered layer. Instead of optimizing
this binary valued array with a Gumbel-Softmax \cite{jang_2016} distribution, the
authors do not learn these values directly. Rather, the method learns a real-valued
matrix $P$ of size $N \times N$, with values in the range $[0, 1]$, where the $(i,
j)$-th element of $P$ represents the proportion of feature channels which are shared by
both the models for task $i$ and task $j$. In this way, the relationships between tasks
are learned directly, and an array $M$ which satisfies $P = \frac{1}{C}M^TM$ is sampled
after each new value of $P$ is computed. With this parameterization of $M$, the network
architecture isn't directly learned, but is instead sampled so that the learned task
affinity matrix dictates the amount of overlap between task parameters. This task
affinity matrix, $P$, is learned through evoluationary strategies. \cite{bragman_2019}
proposes Stochastic Filter Groups (SFGs), in which the assignment of a convolutional
filter to task-specific or shared is learned through variational inference. More
specifically, SFGs are trained by learning the posterior distribution over the possible
assignment of convolutional filters to task-specific or shared roles. As far as we know,
SFGs are the only probabilistic approach to multi-task architecture learning.

\begin{figure}[tb]
    \centerline{\includegraphics[scale=0.2]{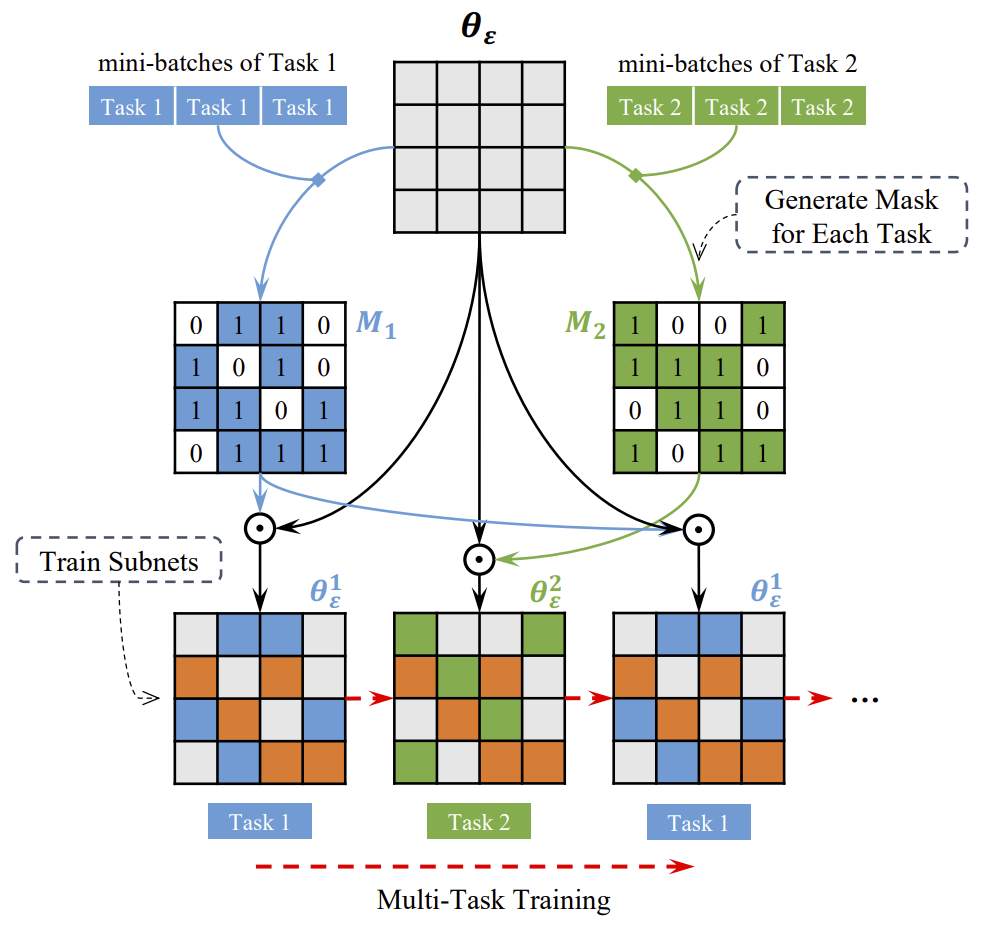}}
    \caption{Learned fine-grained sharing architecture from \cite{sun_2019b}. Each task
    has a sparse subnetwork which may or may not overlap with that of other tasks. Each
    subnetwork is extracted using Iterative Magnitude Pruning \cite{frankle_2018} on the
    entire randomly initialized network before training.}
    \label{sun_2019b}
\end{figure}

\cite{sun_2019b} introduces an algorithm for learning a fine-grained parameter sharing
scheme by extracting sparse subnetworks of a single fully shared model. From a randomly
initialized, overparameterized network, the authors employ Iterative Magnitude Pruning
(IMP) \cite{frankle_2018} to extract a sparse subnetwork from the larger network for
each individual task.  IMP prunes a network by training for a small number of epochs,
then removing the weights which have the smallest magnitude until a desired level of
sparsity is reached. Given a reasonable level of sparsity, the extracted subnetworks for
each task will overlap and exhibit fine-grained parameter sharing between tasks. A
diagram is shown in figure \ref{sun_2019b}. It is important to note that the degree of
overlap between the extracted subnetworks of two tasks is not necessarily correlated
with the relatedness of those two tasks, which suggests the need for a fine-grained
parameter sharing scheme which incorporates information of task affinity to construct
appropriate sharing mechanisms between tasks.

\subsection{Conditional Architectures} \label{conditional_architectures}
Conditional or adaptive computation \cite{bengio_2013} is a method in which parts of a
neural network architecture are selected for execution depending on the input to the
network. Conditional computation is used in many areas outside of multi-task learning,
such as to decrease model computational cost and in hierarchical reinforcement learning
\cite{kulkarni_2016}. In the multi-task case, a conditional architecture is dynamic
between inputs as well as between tasks, though the components of these dynamically
instantiated architectures are shared, which encourages these components to be
generalizable between various inputs and tasks.

\begin{figure}[tb]
    \centerline{\includegraphics[scale=0.225]{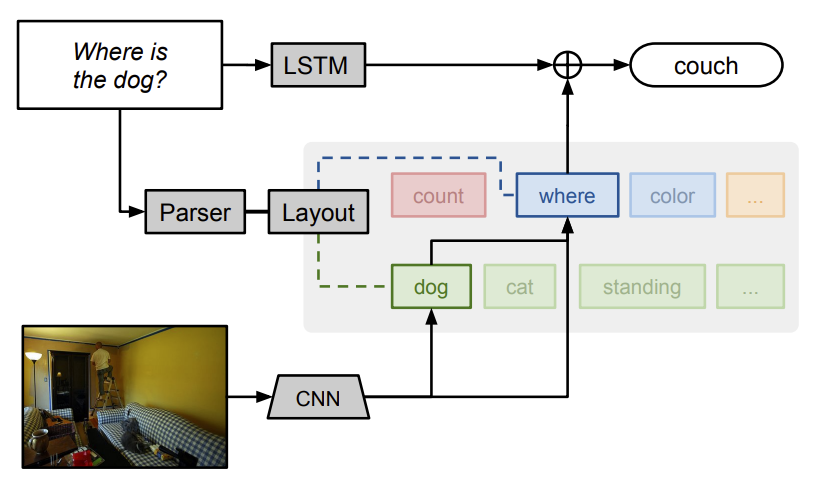}}
    \caption{Example Neural Module Network execution \cite{andreas_2016}. The semantic
    structure of a given question is used to dynamically instantiate a network made of
    modules that correspond to the elements of the question.}
    \label{neural_module}
\end{figure}

Neural Module Networks \cite{andreas_2016} are an early work of conditional computation
which were specifically designed for visual question answering. This method leverages
the compositional structure of questions in natural language to train and deploy modules
specifically catered for the individual parts of a question. The structure of a given
question is determined by a non-neural semantic parser, specifically the Stanford Parser
\cite{klein_2003}. The output of the parser is used to determine the compositional
pieces of the question and the relationships between them, and the corresponding neural
modules are used to dynamically instantiate a model for the given question. This process
is shown in figure \ref{neural_module}. While this work paved the way for future methods
of conditional computation, it is lakcing in the sense that the composition of modules
is not learned. Therefore, the role of each module and combination of modules is fixed
and cannot be improved.

\begin{figure}[tb]
    \centerline{\includegraphics[scale=0.3]{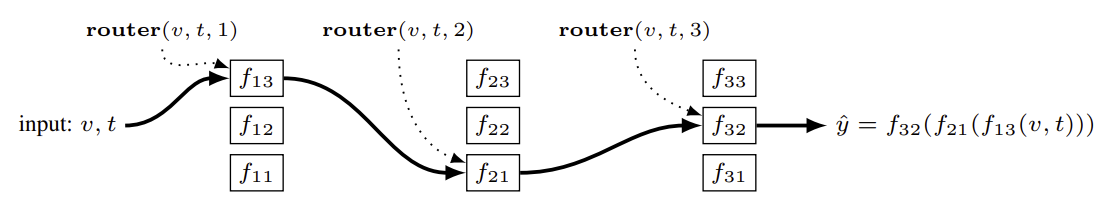}}
    \caption{Example Routing Network execution \cite{rosenbaum_2017}. The router
    iteratively chooses a layer to apply to the input to dynamically instantiate a
    network for each input.}
    \label{routing}
\end{figure}

Routing Networks \cite{rosenbaum_2017} and the Compositional Recursive Learner (CRL)
\cite{chang_2018} are more recent related works of conditional computation in which the
composition of modules is learned in addition to the weights of the modules themselves.
A Routing Network is comprised of a router and a set of neural network modules. Given a
piece of input data, the router iteratively chooses a module from the set of network
modules to apply to the input for a fixed number of iterations; this process is shown in
figure \ref{routing}. The router can also choose a ``pass" action instead of a module,
which simply continues to the next iteration of routing. The module weights can be
learned directly through backpropagation, and the router weights are learned with
reinforcement learning to maximize the performance of the dynamically instantiated
networks on their inputs. The Compositional Recursive Learner of \cite{chang_2018} is
similar, though with some key differences. Given a piece of input data, the CRL also
iteratively chooses a network module from a fixed set of modules through which to route
the input. In the case of the CRL, any task specific information (such as a task ID) is
intentionally hidden from the network modules, to ensure that the modules learn
task-agnostic and therefore generalizable information. CRL is also trained with
reinforcement learning on a curriculum, to encourage the re-use of modules learned on
easier problems.

\cite{ahn_2019} introduces a very similar architecture in which layers of varying
configuration and scale are chosen from a larger backbone network through which to route
the input. The router (called the selector network in this variant) is again trained
with reinforcement learning.

The architecture of \cite{kirsch_2018} is similarly inspired to Routing Networks and the
CRL, but takes a local view of routing rather than a global one. In Routing Networks and
the CRL, any of the network modules can be placed into an instantiated network at any
depth. In contrast, \cite{kirsch_2018} proposes a conditional architecture in which
routing decisions are made only within layers of the network. The architecture is made
of a series of modular layers, each having $m$ network modules. When a layer is to be
applied to an input, the input is passed through a controller, which selects $k$ modules
from the set of $m$ modules belonging to the layer. The layer input is then individually
passed through each of the $k$ selected modules, and the results are added or
concatenated to form the output of the layer. The controllers in these modular layers
are trained not with reinforcement learning but with variational methods, where the
module choice is treated as a latent variable. The authors argue that the architectural
differences in their model from past works on conditional computation diminish the
occurence of \textit{module collapse}, a well-known weakness of conditional models. When
module collapse occurs, the router selects only a small number of modules from the
available set, while the remaining modules remain mostly unused, and the resulting
models do not exhibit modularity.

Most recently, Soft Modularization \cite{yang_2020} is another conditional approach,
which can be seen as a soft relaxation of Routing Networks. Soft Modularization uses
both a router network and a policy network composed of $L$ layers, each with $m$
modules. Instead of making a discrete decision and choosing one module at each step of
computation, as Routing Networks do, the input to each module is a linear combination of
the outputs of modules from the previous layer. Specifically, the router network takes
as input an observation and the corresponding task index, and outputs an $m \times m$
matrix of linear combination weights for each layer after the first, so that the element
in the $i$-th row and $j$-th column of the weight matrix for layer $\ell$ denotes the
weight of module $i$ from layer $\ell - 1$ in the input to module $j$ from layer $\ell$.
The soft relaxation from Routing Networks eliminates the need to train the router
separately from the policy, and instead the entire network can be trained end-to-end.
This architecture is also related to Soft Layer Ordering \cite{meyerson_2017} (see
section \ref{modular_policies}), though with Soft Modularization the linear combination
weights aren't directly learned, instead they are dynamically computed by a separate
network (the router network) at each step of computation. When combined with Soft
Actor-Critic \cite{haarnoja_2018}, the Soft Modularization agent reaches 60\% success
rate on MT50 from the Meta-World benchmark \cite{yu_2019}.

A thorough discussion of the strengths and weaknesses of routing based approaches can be
found in \cite{rosenbaum_2019} and \cite{ramachandran_2018}.

\section{Optimization for Multi-Task Learning} \label{optimization}
With MTL architecture design as the modern generalization of hard parameter sharing on
one side, MTL optimization is the broader version of soft parameter sharing on the
other. Soft parameter sharing is a way to regularize model parameters by penalizing the
distance from model parameters to corresponding parameters of a model for a different,
but related task.  While MTL optimization methods do include regularization strategies
that penalize parameter distance, many other regularization strategies are being
actively developed. When the challenge of negative transfer is viewed through an
optimization lens, new methods for dealing with negative transfer - aside from various
parameter sharing schemes - begin to appear.

We partition the existing MTL optimization methods into six distinct groups: loss
weighting, regularization, gradient modulation, task scheduling, multi-objective
optimization, and knowledge distillation. Just as in previous sections of this review,
the boundaries between these groups are not always concrete. Certain methods may be
interpreted as existing in more than one of these groups, but we believe that this
partition is useful for conceptualizing the various directions of research in MTL
optimization.

\subsection{Loss Weighting} \label{loss_weighting}
A very common approach to ease multi-task optimization is to balance the individual loss
functions for different tasks. When a model is to be trained on more than one task, the
various task-specific loss functions must be combined into a single aggregated loss
function which the model is trained to minimize. A natural question to ask then, is how
to exactly to combine multiple loss functions into one that is suitable for MTL. Most of
the methods we describe here parameterize the aggregated loss function as a weighted sum
of the task-specific loss functions, and the contribution of each method is in the
computation these weights. \cite{gong_2019} contains an empirical comparison of
existing loss weighting methods.

It should be noted that there are several related works \cite{xu_2018a, du_2018} which
introduce methods for weighting the loss of auxiliary tasks relative to a main task
loss. While these methods are interesting and potentially useful for MTL, they were
designed for a setting that lies outside MTL, namely one in which there is a main task
accompanied by one or more auxiliary tasks.

\subsubsection{Weighting by Uncertainty} \label{weight_uncertainty}
One of the earliest methods for learning loss weights is \cite{kendall_2017}. In this
work, the authors treat the multi-task network as a probabilistic model, and derive a
weighted multi-task loss function by maximizing the likelihood of the ground truth
output. For the case of training on $N$ simultaneous regression tasks, the distribution
computed by the network output for task $i$ is the Gaussian $\mathcal{N}(f_i(x),
\sigma_i^2)$, where $f_i(x)$ is the network output for task $i$ and $\sigma_i$ is a
learned parameter which signifies the task-dependent (or homoscedastic) uncertainty for
task $i$. The resulting loss function to jointly maximize the likelihood of each such
distribution is
\begin{equation*}
\sum_{i} \frac{1}{2\sigma^2}\|y_i - f_i(x)\|^2 + \text{log}~\sigma_i
\end{equation*}
Here $y_i$ is the ground-truth label for task $i$. From this derived loss function, we
can see that each task's loss is weighted by the inverse of it's task-dependent
uncertainty, so that tasks with less uncertainty will be given more weight. Also, each
task's loss is regularized by $\text{log}~\sigma_i$, so that the optimization process
isn't incentivized to follow the degenerate strategy of increasing the $\sigma_i$'s
indefinitely. A very similar formula arises for training on classification tasks, and
this method for weighting task losses is the main contribution of this work.
Interestingly, the models trained with this method of weighting were shown to outperform
identical models which were trained with the best performing constant loss weights.

\subsubsection{Weighting by Learning Speed} \label{weight_speed}
Following \cite{kendall_2017}, several methods of weighting multi-task loss functions
were introduced which weigh a task's loss by the learning speed on that task
\cite{chen_2017, liu_2019a, zheng_2018, liu_2019c}, with slightly varying approaches.
The majority of these methods increase a task's loss weight when the learning speed
for that task is low, in order to balance learning between tasks, though this is not the
case for all methods discussed here.

\cite{liu_2019a} and \cite{liu_2019c} each explicitly set a task's loss weight using a
ratio of the current loss to a previous loss. Let $\mathcal{L}_i(t)$ be the loss for
task $i$ at timestep $t$, and let $N$ be the number of tasks. Dynamic Weight Averaging
\cite{liu_2019a} sets the task weights in the following way:
\begin{equation*}
\lambda_i(t) = \frac{N~\text{exp}(r_i(t-1) / T)}{\sum_j \text{exp}(r_j(t-1) / T)}
\end{equation*}
where $r_i(t - 1) = \mathcal{L}_i(t-1) / \mathcal{L}_i(t-2)$ and $T$ is a temperature
hyperparameter. In other words, the loss weight vector is a softmax over the ratios of
successive loss values over the last two training steps for each task, multipled by the
number of tasks. Similarly, Loss Balanced Task Weighting \cite{liu_2019c} sets
\begin{equation*}
\lambda_i(t) = \left( \frac{\mathcal{L}_i(t)}{\mathcal{L}_i(0)} \right)^{\alpha}
\end{equation*}
where $\alpha$ is a hyperparameter. Notice that LBTW measures learning speed as the
ratio of the current loss to the initial loss, while DWA measures it as the ratio of the
losses from the last two training steps. LBTW also does not normalize the weight values
to sum to a fixed value.

GradNorm \cite{chen_2017} is similarly inspired to these two methods, but doesn't
compute loss weights explicitly. Instead, the weights are optimized to minimize an
auxiliary loss which measures the difference between each task's gradient and a desired
task gradient based on the average task loss gradient and the learning speed of each
task. To define this auxiliary loss, we first must define $G_i(t) = \| \nabla_{\theta}
\lambda_i(t) \mathcal{L}_i(t) \|_2$ (weighted gradient for task $i$), $\bar{G}(t)$ as
the average of all such $G_i(t)$, $\tilde{\mathcal{L}}_i(t) = \mathcal{L}_i(t) /
\mathcal{L}_i(0)$ (learning speed for task $k$), and $r_i(t) = \tilde{\mathcal{L}}_i(t)
/ \mathbb{E}_j[\tilde{\mathcal{L}}_j(t)]$ (relative learning speed for task $i$). Then
the auxiliary loss is defined as
\begin{equation*}
\mathcal{L}_{\text{grad}}(\lambda(t)) = \sum_j \|G_j(t) - \bar{G}(t) \times
[r_i(t)]^{\alpha} \|_1
\end{equation*}
where $\alpha$ is again a hyperparameter. By optimizing the task weights $\lambda_i(t)$
to minimize $\mathcal{L}_{\text{grad}}$, the weights are shifted so that tasks with a
higher learning speed yield gradients with smaller magnitude, and tasks with a lower
learning speed yield gradients with a larger magnitude. It should be noted that this
separate optimization adds some compute cost, though the authors only apply GradNorm to
the last layer of shared weights in the network in order to minimize the added cost.
Even with this restriction, GradNorm outperforms baselines.

Notice that all of the methods introduced so far in \ref{weight_speed} increase the
weight of a given task's loss when learning on that task is slower than other tasks. In
comparison, \cite{zheng_2018} assigns a loss weight to a task which decreases as
learning speed increases, assigning a weight of zero if the loss increased on the
previous training step. More specifically, the weight for task $i$ on timestep $t$ is
defined in the following way: Let $\mathcal{L}_i(t)$ be the loss from task $i$ on
timestep $t$, let $\tilde{\mathcal{L}}_i(t) = \alpha \mathcal{L}_i(t) + (1 - \alpha)
\tilde{\mathcal{L}}_i(t-1)$, and $p_i(t) = \text{min}(\tilde{\mathcal{L}}_i(t),
\tilde{\mathcal{L}}_i(t-1)) / \tilde{\mathcal{L}}_i(t-1)$ where $\alpha$ is a
hyperparameter. Then the weight for task $i$ on timestep $t$ is set to
\begin{equation*}
\lambda_i(t) = -(1 - p_i(t))^{\gamma} \text{log}(p_i(t))
\end{equation*}
similar to the Focal Loss \cite{lin_2017}. The rationale behind this strategy is that if
the loss for task $i$ has increased (i.e. $p_i(t) = 1$), there may be a local minimum in
the loss landscape of that task. By assigning a weight for this task to zero, training
steps will only depend on gradients from tasks whose loss is still decreasing, and
gradient descent will (hopefully) escape from the task-specific local minimum in the
landscape of the task whose loss has just increased.

\subsubsection{Weighting by Performance} \label{weight_performance}
Weighting task's losses by performance is similar to weighting by learning speed. These
two categories are distinguished by the fact that learning speed can be thought of as
the rate of change of performance. Given that there are numerous works which introduced
methods for weighting by learning speed, there are surprisingly few methods for
weighting by performance. To our knowledge, the only such works are Dynamic Task
Prioritization \cite{guo_2018} and the implicit schedule in \cite{jean_2019}.

Dynamic Task Prioritization \cite{guo_2018} was inspired by the non-neural MTL work
Self-Paced Multi-Task Learning \cite{li_2016}. Dynamic Task Prioritization (or DTP)
prioritizes difficult tasks and examples by assigning weights both at the task level and
the example level. DTP employs the Focal Loss \cite{lin_2017} to weigh examples within a
task and performance metrics such as classification accuracy to weigh tasks themselves,
where both the example and task level of weights emphasize difficult data over easy
data. These are the distinguishing factors of this work: usage of performance metrics
other than the loss function to weigh tasks, and loss weighting at both the example and
the task level.

The method for loss weighting introduced in \cite{jean_2019} is deemed an implicit task
schedule, in reference to the connection between loss weighting and task scheduling (see
section \ref{task_scheduling}). In this work, the $i$-th task is assigned the weight
\begin{equation*}
\lambda_i = 1 + (\text{sign}(\bar{S} - S_i)) ~ \text{min}(\gamma, (\text{max}_j ~
S_j)^{\alpha} |\bar{S} - S_i|^{\beta})
\end{equation*}
where $S_i$ is the ratio of the current validation performance to a target validation
performance, $\bar{S}$ is the average over all $S_i$'s, $\gamma$ is a hyperparameter
which limits the difference between task weights, $\alpha$ is a hyperparameter that
adjusts how quickly the weights deviate from uniformity, and $\beta$ is a hyperparameter
that adjusts the emphasis on deviations of a task's score from the mean score. While the
formula to compute the loss weights in this implicit schedule looks quite different from
the focal loss, they have the same intention: focus on tasks with poor performance.
Interestingly, this work also includes a discussion of the difference between scaling
learning rates and scaling gradients (there is no difference for vanilla SGD), which is
an often overlooked but important detail of choosing loss coefficients.

\subsubsection{Weighting by Reward Magnitude} \label{weight_magnitude}
It is a well known issue in multi-task learning that a difference in the scale of loss
functions between tasks can cause imbalanced learning dynamics when training jointly on
such tasks. For example, consider an MTL setting with two tasks, $T_1$ and $T_2$, both
classification tasks. Suppose that the loss $\mathcal{L}_1$ for task $T_1$ is a standard
cross-entropy loss, and the loss $\mathcal{L}_2$ for $T_2$ is equal to the standard
cross-entropy loss multiplied by a constant factor of 1000. It is clear in this case
that the gradient for the joint task loss $\mathcal{L} = \mathcal{L}_1 + \mathcal{L}_2$
will be mostly dependent on the network's performance on $T_2$ and very little on that
of $T_1$, so that the multi-task learning will actually be focused mostly on $T_2$.
While this is a somewhat contrived example, the same principle applies to MTL settings
in the wild, where the scale of loss functions may differ greatly. One approach to
tackle this issue is to compute task loss weights based on the magnitude of each task's
loss function.

\cite{hessel_2018} uses PopArt normalization \cite{van_2016} to perform loss weighting
for multi-task deep reinforcement learning. The authors derive a scale-invariant update
rule for actor-critic methods, then extend it to a multi-task setting. The main idea is
to keep a running estimate of the mean and standard deviation of the return from each
timestep, then replace the returns with the normalized versions. The REINFORCE
\cite{williams_1992} algorithm uses an update rule in which the gradient of the
objective with respect to the policy parameter is
\begin{equation*}
(R(t) - v(s_t)) \nabla_{\theta} \text{log}~\pi(a_t|s_t)
\end{equation*}
where $R(t)$ is the return from step $t$, $v$ is the value function, $\theta$ are the
parameters of the policy $\pi$, and $s_t$ and $a_t$ are the state and action at step
$t$. This work replaces this update rule with
\begin{equation*}
\left( \frac{R_i(t) - \mu_i}{\sigma_i} - \tilde{v}_i(s_t) \right) \nabla_{\theta}
\text{log} ~ \pi(a_t|s_t)
\end{equation*}
where $R_i(t)$ is the return on step $t$ for task $i$, $\mu_i$ and $\sigma_i$ are
running estimates of the mean and standard deviation of $R_i(t)$, and $\tilde{v}_i$ is a
normalized value function for task $i$. A similar replacement is made for the update
rule of the value function, and the details can be found in \cite{hessel_2018}. This
normalization constrains the reward function from each task to have a similar
contribution to the update, so that the relative importance of each task is agnostic to
the scale of the reward functions. Applying PopArt to multi-task deep reinforcement
learning significantly improves performance of agents trained with IMPALA
\cite{espeholt_2018} on the DeepMind Lab \cite{beattie_2016} collection of tasks.

\subsubsection{Geometric Mean of Losses} \label{weight_geometric}
While most MTL methods model the network loss as a weighted average of individual task
losses, \cite{chennupati_2019} proposes to compute the geometric mean of task losses as
an alternative. The authors claim that using a geometric mean facilitates balanced
training of all tasks, and that this loss function handles differences in learning
speeds of various tasks better than the traditional weighted average loss function.
However, there is no rigorous evidence to support these claims. The results presented in
their work show that models trained with the geometric outperform baselines, but there
has been little work done on analyzing the specific properties of optimization using
these loss functions in the MTL setting, which may be an interesting direction for
future research.

\subsection{Regularization} \label{regularization}
Regularization has long played in important role in multi-task learning, mostly in the
form of soft parameter sharing. Soft parameter sharing is one of two popular techniques
for MTL (the other being hard parameter sharing) in which parameters aren't shared
between task models, but instead the $L_2$ distance between the parameters of task
models is added to the training objective, in order to encourage similar model
parameters between different tasks. Soft parameter sharing is simple to implement and
has been employed extensively in MTL methods.

\cite{duong_2015} is a well-known method which uses soft parameter sharing instead of
hard parameter sharing. The authors employ the architecture of \cite{chen_2014} for
dependency parsing in multiple languages, but train separate copies of the same network,
one for each language. Only a small fraction of the parameters in the network are softly
shared between tasks, namely the layers which transform the embedded POS tags and the
embedded arc labels. The use of soft parameter sharing across models for different
languages was shown to greatly increase performance in the small data setting.

An interesting variant of soft parameter sharing is introduced in \cite{yang_2016a}, in
which the $L_2$ distance between parameter vectors is replaced by the tensor trace norm
of the tensor formed by stacking corresponding parameter vectors from different tasks.
The trace norm of a matrix is the sum of the singular values of that matrix, and it can
be thought of as a convex relaxation of rank, i.e. the number of non-zero singular
values. Therefore, minimizing the trace norm of a matrix is a good surrogate for
minimizing the rank of that matrix. By extending the trace norm from matrices to tensors
and minimizing the tensor trace norm of stacked parameter vectors, this method
encourages the learning of parameter vectors across tasks which are similar, just as in
traditional soft parameter sharing. In this case, however, similarity is measured by the
existence of linear dependencies between parameter vectors (i.e. low tensor rank),
instead of $L_2$ distance. The authors interpret the resulting trace norms after
training as a measure of \textit{sharing strength} between corresponding layers in
different task models (low trace norm means stronger sharing), and interestingly, the
sharing strength was found to decrease with layer depth. This conincides with the common
intuition in MTL that representations in earlier layers should be less task-dependent
than those of deeper layers.

Besides soft parameter sharing, MTL models can also be regularized by placing prior
distributions on the network parameters. Multilinear Relationship Networks (MRNs)
\cite{long_2017} do exactly this by imposing a tensor normal distribution as a prior
over the parameters in task-specific layers of multi-task models. A tensor normal
distribution is essentially a multivariate normal distribution with the extra assumption
that the covariance matrix can be decomposed into the Kronecker product of $K$
covariance matrices, where $K$ is the order of a tensor following this distribution.
Each of these covariance matrices represents the covariance between rows of various
matricizations of a tensor following the distribution. To impose this distribution on
the parameter tensor of a multi-task network, the covariance is constructed as the
Kronecker product of three covariance matrices: a covariance matrix representing the
relationships between features, another indicating the relationships between
classification classes, and the last modeling the relationships between tasks.  It is
this construction that allows the model to learn relationships between tasks, as its
name suggests. At the time of its publication, MRNs reached state of the art performance
on three different MTL benchmarks.

Deep Asymmetric Multitask Feature Learning (Deep-AMTFL) \cite{lee_2018} is a method of
regularizing deep multi-task neural networks by introducing an autoencoder term to the
objective function. This auxiliary task involves reconstructing the features from the
second to last layer of a network from the network output, so that each of the task
predictions is used to construct the features for all other tasks, a task which was
proposed by Asymmetric Multi-Task Learning \cite{lee_2016}. The motivation behind these
methods is to allow for information to flow from tasks which the model does well to
tasks which the model does poorly, but not the other way around, hence the ``asymmetric"
in the names.

AdaShare \cite{sun_2019a} (architecture discussed in section \ref{modular_sharing})
introduces a novel regularization scheme for MTL methods by regularizing sharing
parameters instead of module parameters. AdaShare uses a set of neural network blocks
which are shared between many tasks, though not all blocks are used by every task. The
sharing parameters of this architecture encode the usage of blocks by different tasks,
as shown in figure \ref{adashare}. AdaShare regularizes these sharing parameters
$\alpha_i$ instead of the network weights. Specifically, the training objective includes
two auxiliary terms
\begin{equation*}
\mathcal{L}_{sparsity} = \sum_{\ell \leq L, i \leq N} \text{log}~\alpha_{i,\ell}
\qquad
\mathcal{L}_{sharing} = \sum_{i, j \leq N} \sum_{\ell \leq L} \frac{L - \ell}{L} \|
\alpha_{i, \ell} - \alpha_{j, \ell} \|_1
\end{equation*}
where $N$ is the number of tasks and $L$ is the number of blocks.
$\mathcal{L}_{sparsity}$ encourages each task's network to be sparse, while
$\mathcal{L}_{sharing}$ encourages similarity in the sharing parameters of different
tasks. Notice that the coefficient $\frac{L - \ell}{L}$ in the definition of
$\mathcal{L}_{sharing}$ linearly decreases the importance of sharing in deeper layers,
which follows the observation in \cite{yang_2016a} that more sharing should occur in
earlier layers, though in this case it is explicitly encouraged. Both of these
regularization terms are very general, and could potentially be applied to Soft Layer
Ordering \cite{meyerson_2017}, Modular Meta Learning \cite{alet_2018}, Stochastic Filter
Groups \cite{bragman_2019}, and many other architectures which learn what to share
between tasks.

A related type of regularization is introduced for conditional computation models,
specifically Routing Networks (discussed in section \ref{conditional_architectures}), in
\cite{cases_2019}. In this case, the regularizer is shaping the decisions of module
selection between tasks by encouraging diversity of choices made by the router, but it
takes a slightly different form compared to the AdaShare regularization, due to the
differences between Routing Networks and the AdaShare architecture. Specifically,
Routing Networks iteratively construct a network layer by layer, choosing between the
set of all layers at each step, so that any permutation (with repetitions) of layers can
be combined into a network by the router. In particular, this means that the router can
ignore many or most layers and only utilize a few layers per task, which is a well known
occurence in training Routing Networks called \textit{module collapse}
\cite{rosenbaum_2019}. Module collapse causes a waste of network components as well as a
decrease in modularity. The issue of module collapse is addressed by the regularization
technique used in \cite{cases_2019}, which rewards diversity of choices made by the
router, so that no layer is ignored.

Finally, Maximum Roaming \cite{pascal_2020} is a multi-task regularization method that
can be thought of as a variant of Dropout \cite{srivastava_2014} specifically made for
MTL networks. While most MTL methods partition parameters between tasks in either a
fixed or principally learned way, Maximum Roaming randomly varies the parameter
partitioning during training, under the constraint that each parameter must be assigned
to a maximal number of tasks. Despite the unorthodoxy of this idea, it is empirically
demonstrated to be beneficial for performance on the CelebA \cite{liu_2015b}, CityScapes
\cite{cordts_2016}, and NYU-v2 \cite{silberman_2012} datasets. The intuitive explanation
of the benefit of roaming is that it allows for each unit to learn from all tasks
throughout training without the constraint that each parameter is always shared between
all tasks, which is a likely contributor to negative transfer. Intuition aside, this
phenomenon is not yet rigorously understood and further work is certainly needed to
fully utilize the potential gains.

\subsection{Task Scheduling} \label{task_scheduling}
Task scheduling is the process of choosing which task or tasks to train on at each
training step. Most MTL models make this decision in a very simple way, either training
on all tasks at each step or randomly sampling a subset of tasks to train on, though
there is some variation in these simple task schedulers. For example, when training on a
single task at each update step in supervised learning settings, it is common to employ
either uniform task sampling \cite{dong_2015}, where each task has the same probability
of being chosen, or proportional task sampling \cite{sanh_2019}, in which the
probability of choosing a task is proportional to the size of its dataset. Despite the
fact that most methods use these baseline task schedulers, it is a well known fact that
optimized task scheduling can significantly improve model performance
\cite{bengio_2009}. 

It is important to note that the problem of scheduling tasks is strongly tied to the
problem of weighting task losses. To see this, consider an MTL setting with two tasks,
$T_1$ and $T_2$, with corresponding loss functions $\mathcal{L}_1$ and $\mathcal{L}_2$,
and consider two separate training setups for this setting. In the first setup, the
model is trained by minimizing the joint loss $\mathcal{L}_1 + 2 \mathcal{L}_2$, and
each training batch holds an equal amount of data from both tasks. In the second setup,
each training batch either holds data exclusively from $T_1$ or $T_2$, where the chances
of the batch containing $T_1$ data and $T_2$ data are $1/3$ and $2/3$, respectively. If
a batch is from $T_1$, the training step will minimize $\mathcal{L}_1$, and if a batch
is from $T_2$, the step will minimize $\mathcal{L}_2$. It isn't hard to intuit that
these setups will, on average, lead to similar results. The training processes won't be
numerically equivalent, but each setup jointly optimizes for the tasks in a way that
prioritizes $T_2$ twice as much as $T_1$. Loss weighting can be seen as a continuous
relaxation of task scheduling, so that many task schedulers can easily be adapted to
loss weighting methods, and vice versa. However, most works adhere to the conventions of
their subfield and only use one of these two framings: multi-task computer vision
methods frequently use loss weighting \cite{dai_2016, misra_2016, ruder_2019}, while
multi-task NLP methods often employ task scheduling \cite{liu_2015a, luong_2015,
liu_2019b}.

\begin{figure}[tb]
    \centerline{\includegraphics[scale=0.225]{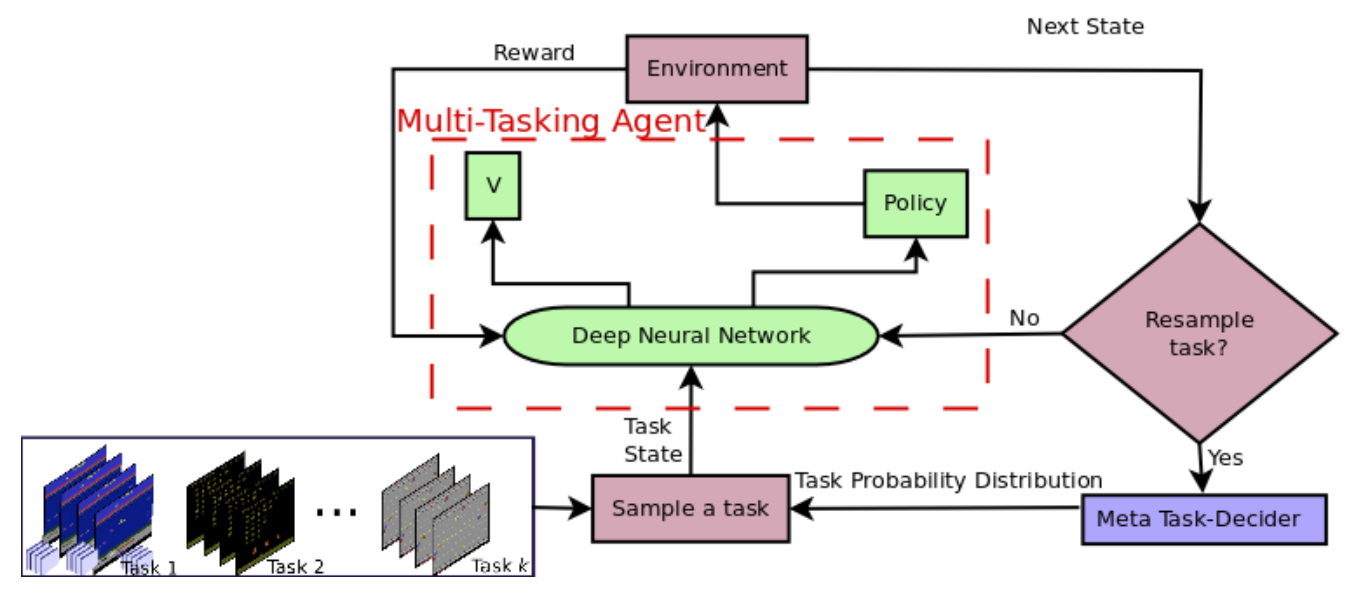}}
    \caption{Task scheduling visualization from \cite{sharma_2017}. A meta task-decider
    is trained to sample tasks with a training signal that encourages tasks with worse
    relative performance to be chosen more frequently.}
    \label{active_sampling}
\end{figure}

\cite{sharma_2017} proposes a method for task scheduling in multi-task RL which is based
on active learning, with three different variants. The common idea behind these three
variants is to assign task scheduling probabilities based on relative performance to a
target level: the further the model is from the target performance on a given task, the
more likely it is that the task will be scheduled. This is akin to the loss weighting
methods that increase the loss weight of a task that exhibits slow learning. Figure
\ref{active_sampling} shows a visualization of the task scheduling process. The
difference between the three variants is in the implementation of the ``meta
task-decider", the component which computes task sampling probabilities. In one way or
another, each variant uses the values $m_i = 1 - a_i/b_i$ to compute these
probabilities, where $i$ is a task index, $b_i$ is the target performance for task $i$,
and $a_i$ is the current model performance for task $i$. Notice that $m_i$ is a measure
of the difference between the current model perfomance and the baseline performance for
task $i$. The first variant, A5C, doesn't learn a sampling distribution, but instead
computes a softmax over all $m_i$'s to construct the sampling distriubtion over tasks.
The second variant, UA4C, treats the task sampling problem as a non-stationary
multi-armed bandit problem in which the reward for the meta task-decider when picking
task $i$ is $m_i$.  This way, the agent is rewarded for choosing tasks which are
furthest from their respective target performances. Lastly, the third variant, EA4C,
treats the sequence of task sampling decisions as a reinforcement learning problem, so
that the meta task-decider can learn to choose sequences of tasks which help the agent
to learn over time. In this case, the reward for the meta task-decider when choosing
task $i$ is
\begin{equation*}
\lambda m_i + (1 - \lambda) \left( \frac{1}{3} \sum_{j \in \mathbb{L}} (1 - m_j) \right)
\end{equation*}
where $\mathbb{L}$ is the task indices of the three tasks with the worst current
performance and $\lambda$ is a hyperparameter. This reward function then incentivizes
the meta task-decider to choose tasks which are furthest from their target performance
while simultaneously choosing tasks which ensure that the performance on the worst tasks
are still improving. Agents trained with these three variants vastly outperform an
identical agent with uniform sampling probability on various collections of Atari games
ranging in size from 6 games to 21 games.

The A5C variant of the algorithm of \cite{sharma_2017} is very similar to a more
recently proposed method for task scheduling \cite{jean_2019}. In this work, each task
is assigned an unnormalized score
\begin{equation*}
\lambda_i = \frac{1}{\text{min}(1, \frac{a_i}{b_i})^{\alpha} + \epsilon}
\end{equation*}
where $a_i$ and $b_i$ are defined similarly as above, and $\alpha$ and $\epsilon$ are
hyperparameters. The unnormalized scores are simply divided by their sums to obtain the
task sampling probabilities. The novel portion of this method is the inclusion of
$\epsilon$ for numerical stability and $\alpha$ to control the degree of over and under
sampling of tasks. \cite{jean_2019} also provides a discussion of task scheduling vs.
loss weighting, in which loss weighting is referred to as ``implicit task scheduling",
as well as a loss weighting method which is discussed in section
\ref{weight_performance}.

\subsection{Gradient Modulation} \label{grad_mod}
One of the main challenges in MTL is \textit{negative transfer}, when the joint training
of tasks hurts learning instead of helping it. From an optimization perspective,
negative transfer manifests as the presence of conflicting task gradients. When two
tasks have gradients which point in opposing directions, following the gradient for one
task will decrease the performance on the other task, and following the average of the
two gradients means that neither task sees the same improvement it would in a
single-task training setting. Among many other approaches to alleviate the conflict in
learning dynamics between different tasks, explicit gradient modulation has arisen as a
potential solution. The methods presented here work by modifying training gradients,
either through the use of adversarial methods or by simply replacing gradient vectors
when conflicts arise.

\subsubsection{Adversarial Gradient Modulation} \label{adv_grad_mod}
\begin{figure}[tb]
    \centerline{\includegraphics[scale=0.25]{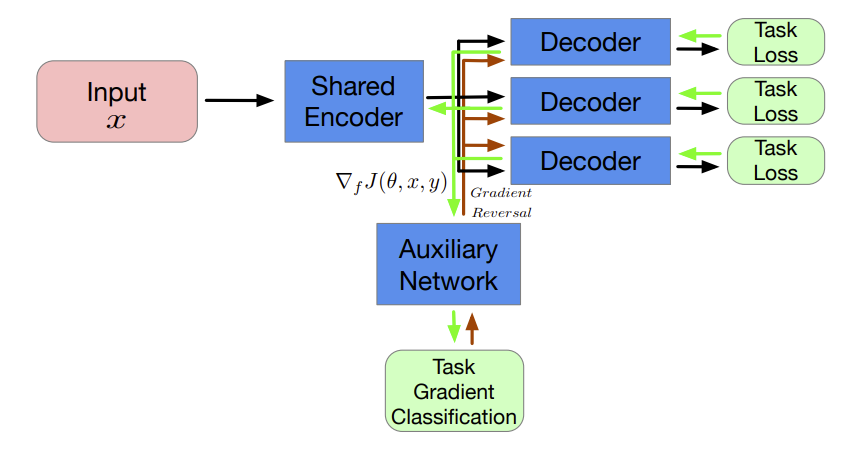}}
    \caption{Multi-task GREAT model \cite{sinha_2018}. An auxiliary network takes a
    gradient vector for a single task's loss and tries to classify which task the
    gradient vector came from. The network gradients are then modulated to minimize the
    performance of the auxiliary network, to enforce the condition that gradients from
    different task functions have statistically indistinguishable distributions.}
    \label{great}
\end{figure}

If a multi-task model is training on a collection of related tasks, then ideally the
gradients from these tasks should point in similar directions. Gradient Adversarial
Training (GREAT) \cite{sinha_2018} explicitly enforces this condition by including an
adversarial loss term that encourages gradients from different sources to have
statistically indistinguishable distributions. GREAT is a general framework which can be
applied for adversarial defense and knowledge distillation (and likely other settings)
besides multi-task learning. In the MTL setup, the model is augmented with an auxiliary
discriminator network which attempts to classify the tasks corresponding to gradients of
the task decoders, as pictured in figure \ref{great}. During the backward pass, the
gradients are modified by Gradient Alignment Layers (GALs) through element-wise scaling
to minimize the performance of the auxiliary network in distinguishing between the task
gradients. A similar adversarial setup to enforce gradient similarity between tasks is
used in \cite{maninis_2019}.

While the motivation for the model is intuitively plausible, the premise of adversarial
training isn't rigorously justified. Just because two tasks are related, how can we be
sure that their gradient distributions should be identical? Furthermore, it seems likely
that the distribution of a task's gradients will change throughout training, so how
likely can it be that each task's gradient distributions move together? The existence of
negative transfer in the first place tells us that similar tasks do not necessarily have
aligning gradients. Can we actually alleviate negative transfer by enforcing gradients
to be similar, even when the original gradients of the loss function are conflicting?
Without theoretical justification, we can't be sure of the answers to these questions.
Nevertheless, the experiments presented in this work show that GREAT does increase the
performance of multi-task models, and that it outperforms other multi-task optimization
baselines such as GradNorm. The nature of both the premise and the results of this model
are still unanswered questions.

\subsubsection{Gradient Replacement}
An entirely different approach to gradient modulation is explored in \cite{lopez_2017,
chaudhry_2018, yu_2020}. The main idea behind these three works is to replace a task
gradient vector which conflicts with another by a modified version which has no
conflicts. This idea is broad, but the implementations of each of these works are
similar at heart and we will present a rigorous definition of each.

\cite{lopez_2017} introduces Gradient Episodic Memory (GEM) for continual learning, a
problem formulation in which a model learns multiple tasks sequentially, instead of
simultaneously as in MTL. GEM keeps an episodic memory of training examples from past
learned tasks, and enforces the following constraint at each update step $t$ when
training on task $i$:
\begin{equation*}
\forall j < i: G_i(t)^T G_j(t) \geq 0
\end{equation*}
where $G_i(t)$ is the gradient vector for task $i$ (the current task) and $G_j(t)$ is
the gradient of the loss on the data in episodic memory from task $j$, at training step
$t$. The condition that the dot product between two gradient vectors is non-negative is
equivalent to the condition that the angle between the two gradient vectors is less than
90 degrees, so that they don't point in opposing directions. If this condition isn't met
for some $j$, then $G_i(t)$ is replaced by $\tilde{G}_i(t)$, the solution to the
following optimization problem:
\begin{align*}
\text{minimize:}~ & \frac{1}{2} \|G_i(t) - \tilde{G}_i(t)\|^2 \\
\text{subject to:}~ & \forall j < i: \tilde{G}_i(t)^T G_j(t) \geq 0
\end{align*}
This quadratic optimization problem can be solved efficiently by instead solving the
dual and recovering the corresponding value of $\tilde{G}_i(t)$. Even so, GEM introduces
significant increase in computation time compared to traditional training. Averaged GEM
(A-GEM) \cite{chaudhry_2018} was proposed to alleviate the computation burden. The
authors point out that it is much more efficient to relax the GEM constraint to
\begin{equation*}
G_i(t)^T G_{\text{avg}}(t) \geq 0
\end{equation*}
where $G_{\text{avg}}(t) = \frac{1}{i-1} \sum_{j < i} G_j(t)$. In other words, instead
of requiring the new gradient to be non-conflicting with the task gradient of each
previous task, A-GEM only requires that the new gradient be non-conflicting with the
average of the previous tasks' gradients. By doing this, the modified optimization
problem has the following closed form solution:
\begin{equation*}
\tilde{G}_i(t) = G_i(t) - \frac{G_i(t)^T G_{\text{avg}}(t)}{G_{\text{avg}}(t)^T
G_{\text{avg}}(t)} G_{\text{avg}}(t)
\end{equation*}
This slight relaxation of the constraints yields a huge improvement in computation time
while maintaining the performance of GEM.

This exact update rule is adapted for the MTL setting in \cite{yu_2020} with a method
named PCGrad. Besides the theoretical analysis in the paper, the PCGrad algorithm itself
is near identical to A-GEM. The main difference is due to the difference in problem
formulations: PCGrad is meant to be used when learning multiple tasks simultaneously, so
multiple gradient vectors must be checked for conflicts with others at each update step.
When combined with Soft Actor-Critic \cite{haarnoja_2018}, PCGrad is able to
successfully complete 70\% of the tasks in the MT50 benchmark of the Meta-World
environment \cite{yu_2019}, a challenging, recently proposed environment for multi-task
and meta-learning with robotic manipulation tasks.

The success of gradient modulation methods demonstrate that minimizing the presence of
conflicting gradients between tasks is an effective way to decrease negative transfer.
Continuing to develop such methods may be an important part of MTL optimization in
future research.

\subsection{Knowledge Distillation}
Originally introduced for compressing large ensembles of non-neural machine learning
models into a single model \cite{bucila_2006}, \textit{knowledge distillation} has found
many applications outside of its originally intended domain. In MTL, the most common use
of knowledge distillation is to instill a single multi-task ``student" network with the
knowledge of many individual single-task ``teacher" networks. Interestingly, the
performance of the student network has been shown to surpass that of the teacher
networks in some domains, making knowledge distillation a desirable method not just for
saving memory, but also for increasing performance.

The first applications of policy distillation for multi-task learning came from two
separate papers at the same time (uploaded to arXiv on the exact same day!), namely
Policy Distillation \cite{rusu_2015} and Actor-Mimic \cite{parisotto_2015}. Both of
these methods are designed for reinforcement learning, and follow roughly the same
template: For each task in a collection of tasks, use reinforcement learning to train a
task specific policy to convergence, and after training, use supervised learning to
train a single student policy to mimic the outputs of the task-specific teacher
policies, such as with a mean-square error or cross-entropy loss. Additionally,
Actor-Mimic includes a feature regression objective, where each teacher network has a
corresponding feature prediction network which attempts to predict the hidden
activations of the teacher network from the hidden activations of the student network.
The gradients of this objective are propagated through the student network, so that the
student network is trained to compute features which contain the same information as
each teacher network. Actor-Mimic was also shown to demonstrate impressive transfer
performance. Transfer to new tasks was performed by removing the last layer of the
distilled student policy and using the weights as the initialization for a single-task
policy. The transferred policies were able to learn some new tasks faster than policies
trained from scratch, though occasionally this transfer would slow down learning on new
tasks. Also, both of these papers show similar results for the student network in the
Atari domain: the distilled student network either matches or outperforms the
single-task teachers. This is somewhat surprising, given that the student network is not
trained to maximize in-game reward, it is only trained to mimic the behavior of the
teacher networks.

A common intuitive explanation for the phenomenon of student networks outperforming
their teachers is that the student networks are provided with a more rich training
signal than the teacher. For example, in classification, each single-task network is
provided with a ground truth label for each input in the form of a one-hot vector.
Meanwhile, the student's training signal will be a ``softer" version of this, namely the
teachers' output, a dense vector which may contain information about similarity of
classes to the ground-truth class and other information not found in the ground truth
one-hot vector.

\begin{figure}[tb]
    \centerline{\includegraphics[scale=0.25]{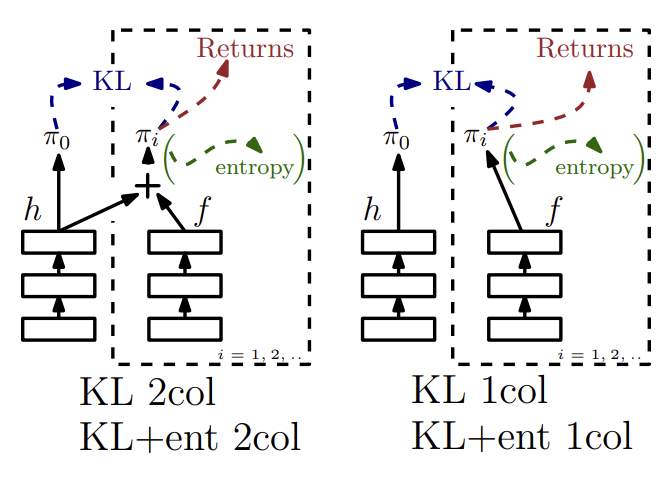}}
    \caption{Two architectures from the Distral framework for RL \cite{teh_2017}. On the
    left is an architecture which employs both of the main ideas behind Distral:
    KL-regularization of single-task policies with the multi-task policy and a
    two-column policy for each task, where one column is shared between all tasks. On
    the right is an architecture which only employs KL-regularization of the single-task
    policies.}
    \label{distral}
\end{figure}

It is interesting to note that most knowledge distillation algorithms (these two
included) have an asymmetric information flow between student and teacher, namely that
information travels from teacher to student, but not the other way around. This
observation raises the question: should the single-task teacher networks receive
information from the distilled multi-task student network? This isn't possible with the
methods discussed so far, since the teacher networks are done training before the
student network starts it. On the other hand, the Distral framework for multi-task
reinforcement learning \cite{teh_2017} provides a setting which accomplishes exactly
this symmetric information flow between student and teacher. Distral is a very general
framework which leads to several different loss functions and architectures, though each
variant is driven by one or both of two main ideas: The single-task policies are
regularized by minimizing the KL-divergence between single-task policies and the shared
multi-task policy as a part of the training objective, and the policies for each task
are formed by adding the output of the corresponding single-task policy with the output
of the shared multi-task policy. Two of the resulting architecture variants are pictured
in figure \ref{distral}. The details of each variation and the motivation behind the
design choices can be found in the original work. It should be noted, though, that the
lines between different approaches to multi-task RL being to blur when considering
Distral. This framework does not use knowledge distillation in the same sense as Policy
Distillation and Actor-Mimic, since the shared multi-task network isn't explicitly
trained to mimic the outputs of the single-task network.

Most recently, knowledge distillation was applied to multi-task NLP with MT-DNN
ensembles \cite{liu_2019d} and Born-Again Multi-tasking networks (BAM)
\cite{clark_2019}. Both works mainly use the original template for multi-task knowledge
distillation, but the authors of BAM also introduce a training trick to help student
networks surpass their teachers which they name teacher annealing. For model input $x$,
ground truth label $y$, and teacher output $f_T(x)$, the usual target output for the
student on a given example $x$ is $f_T(x)$. With teacher annealing, the student's target
output is replaced by $\lambda y + (1 - \lambda) f_T(x)$, where $\lambda$ is linearly
annealed from 0 to 1 throughout student training. This way, by the end of the student
training process, the student is trying to output the ground truth labels for each input
and is no longer trying to mimic the teacher, so that the student isn't inherently
limited by the teacher's weaknesses. Ablation studies in this work show that teacher
annealing does improve student performance on the GLUE benchmark.

\subsection{Multi-Objective Optimization}
The need to optimize for multiple - possibly conflicting - loss functions is a
fundamental difficulty of MTL. The standard formulation of machine learning involves the
optimization of a single loss function, so that the methods created to solve such
problems only consider a single loss function. As we've seen so far, most MTL methods
circumvent this challenge by combining many loss functions into one using a weighted
average, though this fix isn't perfect. The map from a tuple of loss values
$(\mathcal{L}_1(t), \mathcal{L}_2(t), ..., \mathcal{L}_N(t))$ to their weighted average
$\sum_i \lambda_i \mathcal{L}_i(t)$ isn't an injective mapping, meaning that some
information is lost when we transform a collection of loss functions into a single
weighted loss function \footnote{It is a well known fact in mathematical analysis that
there is no continuous injective mapping from $\mathbb{R}^d$ to $\mathbb{R}$ for $d \geq
2$, so unfortunately there is no immediate candidate for a multi-task loss function
which is superior to the weighted average in the sense of injectivity.}. Constructing
this weighted average also necessitates a choice of weights, which is prone to error.
Using multi-objective optimization for MTL is an alternative optimization method which
doesn't suffer from these weaknesses.

\begin{figure}[tb]
    \centerline{\includegraphics[scale=0.175]{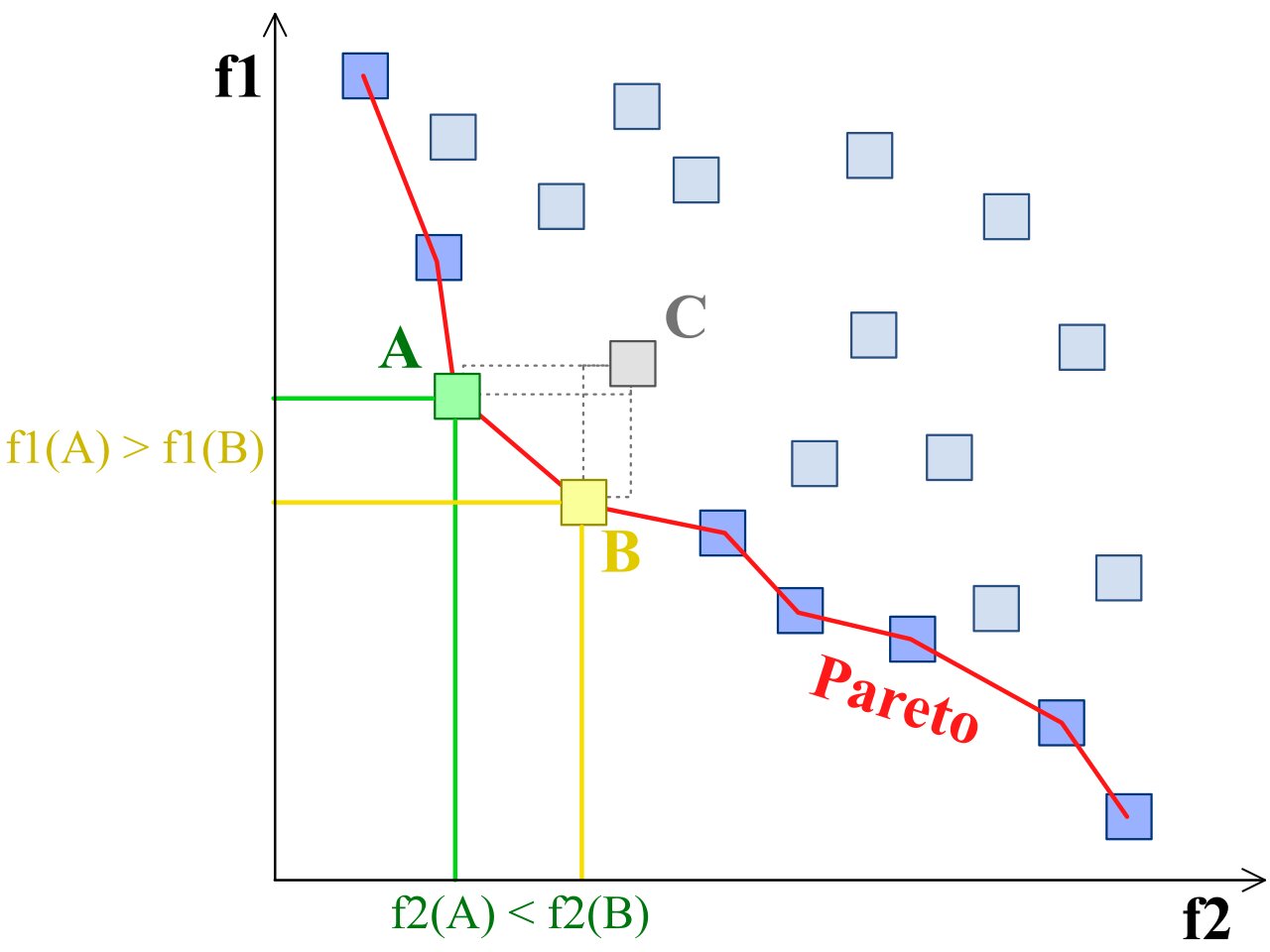}}
    \caption{Visualization of Pareto optimal solutions for a two-objective optimization
    problem \cite{dreo_2006}. The Pareto frontier is made of all points along the red
    curve.}
    \label{pareto_frontier}
\end{figure}

Multi-objective optimization is exactly the process of optimizing several objective
functions simultaneously. Notice that in a multi-objective problem, there is not
necessarily a solution which is a global minimum for all objective functions, meaning
that typically there are no globally optimal solutions for multi-objective optimization
problems. Instead, we consider solutions which are Pareto optimal. Pareto optimal
solutions to a multi-objective problem are those for which the performance for any of
the objectives can only be improved by worsening performance on another objective. In
other words, Pareto optimal solutions represent the best feasible options for solving a
multi-objective optimization problem, up to a trade-off between objectives. The set of
Pareto optimal solutions to a multi-objective optimization problem is called the Pareto
frontier, and is visualized in figure \ref{pareto_frontier}.

Despite being a natural fit for MTL and a well-studied problem \cite{miettinen_1998}, it
has only recently been applied to multi-task problems. \cite{sener_2018} brought
gradient-based multi-objective optimization algorithms to the field of deep multi-task
learning by extending the well known Multiple Gradient Descent Algorithm (MGDA)
\cite{desideri_2012} to a form that scales well to the high dimensionality of deep
learning problems. This is accomplished by minimizing an upper bound to the MGDA loss,
and doing so incurs only a small amount of computational overhead compared to
traditional MTL. Pareto Multi-Task Learning \cite{lin_2019} takes this extension one
step further by generalizing the algorithm proposed in \cite{sener_2018} in order to
compute multiple Pareto optimal solutions. Since no Pareto optimal solution is a priori
superior to any other, a set of Pareto optimal solutions which is representative of the
Pareto frontier is more flexible and likely more useful than a single solution. Pareto
MTL works by decomposing the multi-objective optimization problem into multiple
subproblems, each with varying preferences between objectives. Interestingly, the
authors show that Pareto MTL and the algorithm presented in \cite{sener_2018} can
actually be formulated as methods to compute adaptive loss weights, similar to those
discussed in section \ref{loss_weighting}. This is somewhat counterintuitive, since
multi-objective optimization methods are intended to be of a fundamentally different
nature than methods which optimize a weighted average over loss functions. With further
exploration, the surprising connection between these two directions could potentially
lead to a better understanding of existing multi-task optimization methods.

\section{Task Relationship Learning} \label{relationship}
We have now discussed MTL architectures and optimization methods, completing a broader
analogue of the popular dichotomy specified by hard and soft parameter sharing.
However, there is a lesser known third wheel to this pair of approaches: task
relationship learning. Task relationship learning (or TRL) is a separate approach that
doesn't quite fit into either architecture design or optimization, and is more specific
to MTL. The goal of TRL is to learn an explicit representation of tasks or relationships
between tasks, such as clustering tasks into groups by similarity, and leveraging the
learned task relationships to improve learning on the tasks at hand.

In this section we discuss three research directions within TRL. The first is grouping
tasks, where the goal is to partition a collection of tasks into groups such that
simultaneous training of tasks in a group is beneficial. The second is learning transfer
relationships, which includes methods that attempt to analyze and understand when
transferring knowledge from one task to another is beneficial for learning. Finally, we
discuss task embedding methods, which learn an embedding space for tasks themselves.

\subsection{Grouping Tasks}
As a solution to negative transfer, many MTL methods are designed to adaptively share
information between related tasks and separate information from tasks which might hurt
each other's learning. The papers discussed here use task grouping as an alternative
solution: if two tasks exhibit negative transfer, simply separate their learning from
the start. However, doing so requires significant computation time for trial and error
in training networks jointly for various sets of tasks, and there are currently very few
methods which can accurately determine the joint learning dynamics of groups of tasks
without this kind of brute force trial and error.

Two early concurrent works of learning to group tasks are \cite{alonso_2016} and
\cite{bingel_2017}. Both of these papers are empirical studies analyzing the
effectiveness of various task groupings in MTL for natural language processing, with a
focus on choosing one or two auxiliary tasks (such as POS tagging, syntactic chunking,
and word counting) to help learning on a main task (such as named entity recognition and
semantic frame detection) by training a multi-task network on many combinations of
tasks. In these studies, a single-task network is trained for each individual main task,
and its performance is compared to the performance of a multi-task network trained on
the main task jointly with one or two auxiliary tasks.

\cite{alonso_2016} trains 1440 task combinations, each with a main task and one or two
auxiliary tasks, and finds that performance on the main task improves the most with
auxiliary tasks whose label distributions have high entropy and low kurtosis.  This is
consistent with the findings of \cite{bingel_2017}, in which 90 pairs of tasks (one
main, one auxiliary) are trained. Using the results of these training runs as data, this
work trains a logistic regression model to predict whether an auxiliary task will help
or hurt main task performance based on features from the datasets and learning curves of
the two tasks. They also find that entropy of the auxiliary label distribution is highly
correlated with improvement on the main task, though the features most highly correlated
with main task improvement are the gradients of the main task learning curve when
trained on its own. Specifically, if the learning curve of a task (when trained in a
single-task setup) begins to plateau during the first 20\% to 30\% of training,
including an auxiliary task in training is likely to improve the performance on the main
task. The authors speculate that this may be because a main task whose learning curve
has an early plateau is likely to be stuck in a non-optimal local minimum, and the
inclusion of an auxiliary task helps the optimization process to escape this minimum.
Somewhat surprising is their finding that the difference in sizes of the main and
auxiliary task dataset was not found to be indicative of the main task performance gain
when including the auxiliary task. Despite the fact that these studies don't treat all
tasks identically, as is usually the case with MTL, the conditions they find which imply
positive transfer between tasks are general enough that they may be useful in the
multi-task setting.

An empirical study on the joint training of computer vision tasks is performed in
\cite{doersch_2017}, with a focus on self-supervised tasks, namely relative position
regression, colorization, motion segmentation, and exemplar matching. This study is less
in-depth, as it is not the sole focus of the paper, but the authors come to the
interesting conclusion that multi-task training always improved performance compared to
the single-task baselines. This fact is very surprising given the inconsistency of
improvement that MTL usually affords over single-task training. The consistent
improvement may be due to the relationships between the tasks or the nature of their
self-supervised labels, but none of these answers are certain. 

\begin{figure}[tb]
    \centerline{\includegraphics[scale=0.25]{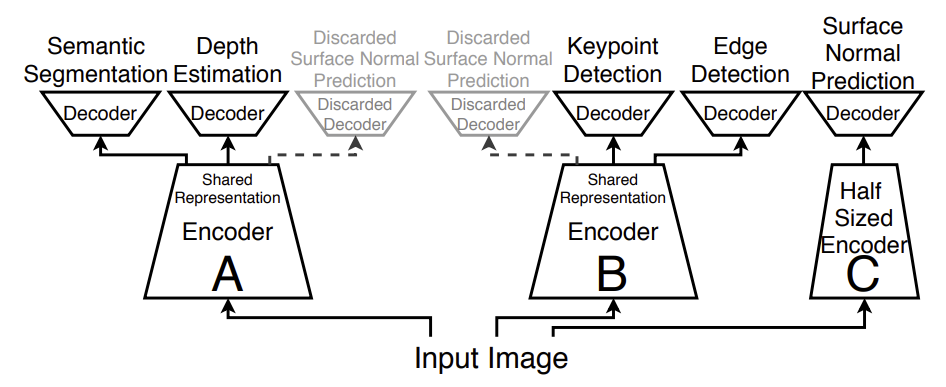}}
    \caption{An example partitioning of a group of tasks into clusters with positive
        transfer \cite{standley_2019}.}
    \label{which_tasks}
\end{figure}

Adjacent to these empirical studies to analyze multi-task task relationships is a
principled method for learning these relationships online during training without
trial-and-error task grouping, called Selective Sharing \cite{strezoski_2019b}.
Selective sharing uses a shared trunk architecture to handle multiple tasks, and
clusters tasks into groups based on the similarity of their gradient vectors throughout
training. This clustering is motivated by the fact that the task-specific branches are
all initialized with identical parameters, so that the similarity between task gradients
is indicative of the similarity of tasks. As the clusters of tasks are updated
throughout training, the task branches of the network are merged so that tasks which are
clustered together share parameters, and this process continues until the clusters stop
changing. Aside from the obvious benefit of decreased computation cost compared to large
scale empirical studies to determine groups of tasks, this method uses learned task
features to understand the relationships between tasks, which is a powerful and
inexpensive approach to TRL that is also employed in \cite{kriegeskorte_2008, song_2019}
(see section \ref{transfer_rel} for further discussion). It should be noted, however,
that their model is based on an assumption which breaks down more and more during
training. It may be true that gradient vectors are indicative of task similarity at the
beginning of training, when parameters across tasks are still relatively similar. But as
training continues and model parameters get further apart, similarity between task
gradients becomes less and less representative of the similarity between tasks, and this
signal will devolve into noise with a sufficiently non-convex loss landscape. Still, the
approach is empirically shown to be effective for computing task relationships with
proper configuration.

Most recently, \cite{standley_2019} includes an in-depth empirical study of task
grouping with the Taskonomy dataset \cite{zamir_2018} and a method to partition a group
of tasks into clusters which each exhibit positive transfer between their respective
tasks. Such a partitioning of tasks is pictured in figure \ref{which_tasks}. Using four
different training settings with varying amounts of training data and network sizes to
train each pair of tasks within groups of five tasks, the authors find several
interesting trends with a more thorough analysis than the previous studies. First, there
were mixed results on whether or not multi-task training improved over the single-task
baselines, with many multi-task networks performing worse than the single-task
counterparts. Next, the performance gain from single-task to multi-task training varies
wildly with the training setting, implying that the effectiveness of MTL is not as
dependent on the relationship of the tasks themselves as we might have once thought.
Surprisingly, the study also finds no correlation between the multi-task affinity and
the transfer affinity between tasks, which again shows that there are many more factors
behind joint task learning dynamics (in both multi-task and transfer learning) than just
the nature of the tasks in consideration. To find a partition of a group of tasks into
clusters with desirable learning dynamics, this work uses both approximations of the
performance of multi-task networks at convergence and a branch-and-bound algorithm that
uses these approximations to select a set of multi-task networks to collectively perform
all tasks. Using this method of grouping tasks, the resulting multi-task networks
consistently outperform the single task baselines, which is a vast improvement over the
multi-task setups from the empirical study in which every single pair of tasks is
trained jointly. To our knowledge, this is the only computational framework for deciding
which tasks to train together in multi-task learning that allows for more than two tasks
to be trained jointly.

\subsection{Transfer Relationships} \label{transfer_rel}
Learning transfer relationships between tasks in MTL is related to the problem of
learning to group tasks for joint learning, though they don't always correlate, as noted
above. However, unlike learning tasks simultaneously, transfer learning already plays an
important role in the wider deep learning research effort; most natural language
processing and computer vision models start not from scratch, but transferring a
pre-trained model to use on a new task. Be that as it may, research into methods that
can explicitly learn transfer relationships between tasks is only somewhat recent. With
the large existing applicability of transfer learning today, these methods have the
potential to make a strong impact on the larger research community.

\begin{figure}[tb]
    \centerline{\includegraphics[scale=0.275]{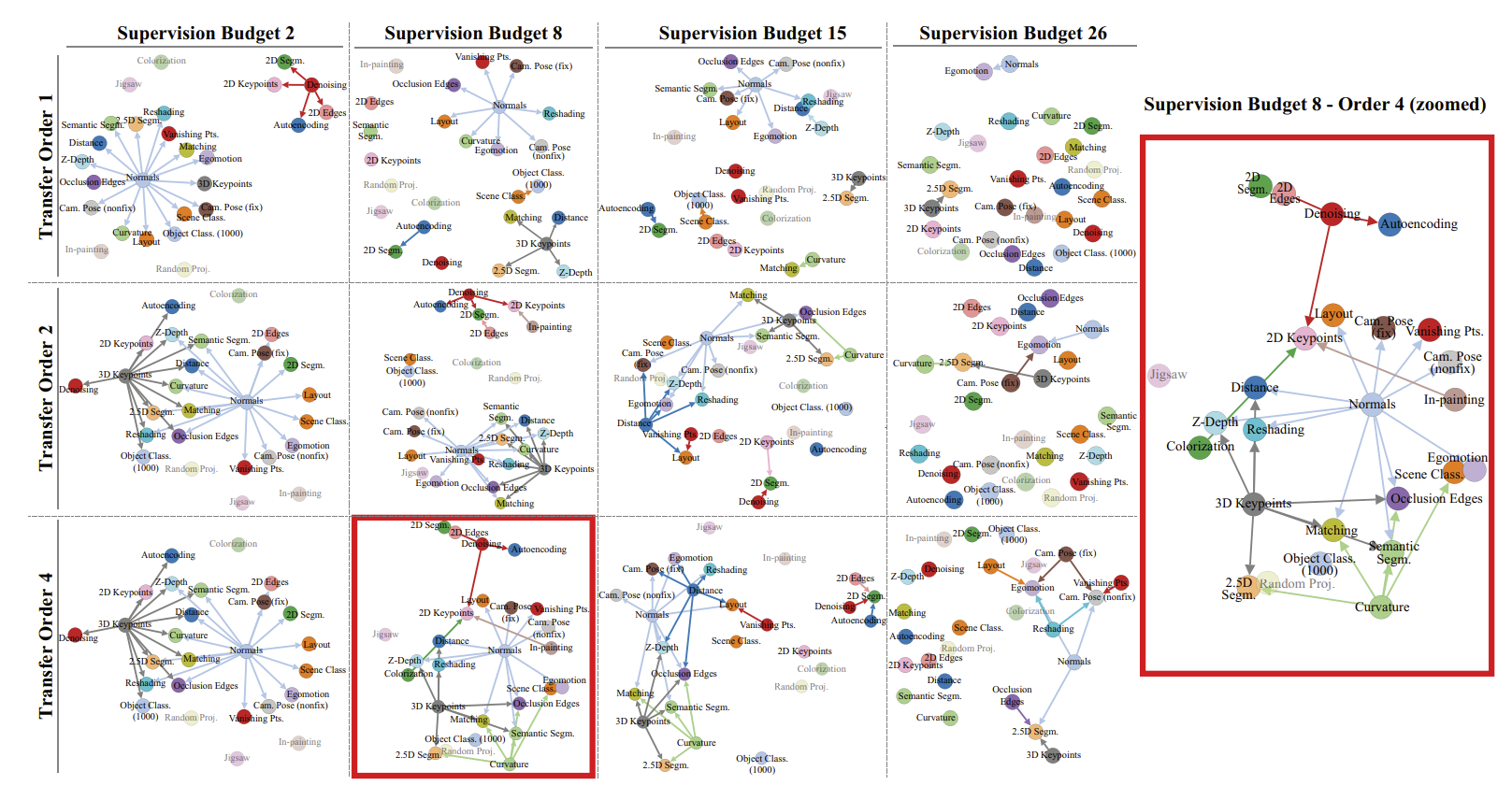}}
    \caption{Task taxonomies for a collection of computer vision tasks as computed in
    Taskonomy \cite{zamir_2018}. An edge from task $i$ to task $j$ denotes that task $i$
    is an ideal source task to perform transfer learning on task $j$.}
    \label{taskonomy}
\end{figure}

The first (and certainly most well known) work which attempted to learn transfer
affinities between tasks is Taskonomy \cite{zamir_2018}. Aside from the Taskonomy
dataset with 4 million images labeled for 26 tasks, this paper introduces a
computational method to automatically construct a taxonomy of visual tasks based on
transfer relationships between tasks. To do this, a single-task network is trained on
each individual task, then transfer relationships are computed by answering the
following question for each pair of tasks: How well can we perform task $i$ by training
a decoder on top of a feature extractor which was trained on task $j$? This is a bit of
a simplification, as the actual training setup involves transferring from multiple
source tasks to a single target task, but the main idea is the same. Once the transfer
affinities are computed, the problem of constructing a task taxonomy is characterized as
choosing the ideal source task or tasks for each target task in a way that satisfies a
budget on the number of source tasks. The motivation here is to limit the number of
tasks which have access to the full amount of supervised data (these are the source
tasks), and to learn the remainder of tasks by transferring from the source tasks, with
only a small amount of training data to train the decoder on top of the transferred
feature extractor. The problem of choosing the ideal set of source tasks and which
source tasks to use for each target task (given the task transfer affinities) is encoded
as a Boolean Integer Programming problem. The solution can be represented as a directed
graph in which the nodes are tasks, and the presence of an edge from task $i$ to task
$j$ means that task $i$ is included in the set of source tasks for task $j$. Some
resulting taxonomies for varying supervision budgets and transfer order (maximum number
of source tasks for each target task) are shown in figure \ref{taskonomy}. Taskonomy is
the first large scale empirical study to analyze task transfer relationships and compute
an explicit hierarchy of tasks based on their transfer relationships, and by doing so
they are able to compute optimal transfer policies for learning a group of related tasks
with limited supervision. However, their method of doing so is extremely expensive,
since it involves training for a huge number of combinations of source/target tasks. The
entire process of constructing task taxonomies took 47,886 GPU hours.

A similarly inspired but much more efficient method for learning task transfer
relationships is introduced in \cite{dwivedi_2019}, which uses Representation Similarity
Analysis (RSA) \cite{kriegeskorte_2008} to compute a measure of similarity between
tasks. RSA is a commonly used tool in computational neuroscience to quantitatively
compare measures of neural activity, and it has been adopted for analyzing neural
network activations by the deep learning community in recent years
\cite{vandenhende_2019}. The underlying assumption behind the RSA transfer model in
\cite{dwivedi_2019} is that if two tasks would exhibit positive transfer, then
single-task networks trained on each of them will tend to learn similar representations,
and so RSA will be an accurate measure of the transfer affinities of the task at hand.
Because RSA only involves comparing the representations of different networks, there is
no need to actually perform any transfer learning between each pair of tasks, making the
RSA transfer model orders of magnitude faster than Taskonomy. Furthermore, the authors
find that the computed task affinities from RSA are nearly independent of the size of
the models used to train on the tasks, so that the computation of the task relationships
can be done with very small models in order to cut computation cost even more.

Most recently, \cite{song_2019} follows a similar approach as the RSA transfer model:
compare the similarity of single-task networks to compute task transfer affinities,
instead of actually performing transfer learning. Instead of comparing the networks'
learned representations, the method presented in \cite{song_2019} compares their
attribution maps on the same input data. An attribution map is a scoring over the
individual units of a network's input which represents the relevance of each unit to the
network's output. In computer vision, for example, an attribution map assigns a
relevance score to each pixel in the input. Just as the RSA transfer model assumes that
tasks with positive transfer will learn similar representations, the attribution map
transfer model assumes that such tasks will pay attention to the same parts of an input.
This approach shows similar results as the RSA transfer model: orders of magnitude
speedup compared to Taskonomy without degradation of the results. Unfortunately, this
work doesn't include any direct comparison with the RSA transfer model, so there is no
evidence of superiority of either model.

The existing methods for learning task transfer relationships are all very recent, and
there is much more work to be done in this area. One interesting thing to note is the
manner in which the RSA and attribution map transfer models \cite{kriegeskorte_2008,
song_2019} achieve efficiency while computing nontrivial information. To summarize
succinctly, these models use the network to train the network. Both methods leverage
information learned by the single-task networks (either intermediate representations or
relevance scoring) in order to inform training downstream. Taskonomy, on the other hand,
trains extra networks to do what these two methods did without any extra training. It
goes to show that the rich information learned by deep networks isn't only useful for
the network's forward pass. In general, even outside of MTL, this information can and
should be leveraged to further inform model training: Use the network to train the
network.

\subsection{Task Embeddings}
Although they are mostly used for meta-learning, task embeddings are a very general form
of learning task relationships, and are strongly related to the methods we have so far
discussed in this section. Even with this strong tie between models, there is a
significant lack of methods in MTL which utilize task embeddings. This shouldn't come as
a surprise, though. Task embeddings find their most use in situations where a new task
is given after already learning a number of tasks from the same distribution, and this
new task must be localized with respect to tasks already learned. If the set of tasks
for a model to learn is fixed - as is the case with MTL - why should one assign a vector
representation to each task? Still, we feel that the connection to TRL is important, so
we provide a brief summary of several task embedding methods in the meta-learning
literature.

\cite{james_2018} uses metric learning to construct a task embedding for imitation
learning of various robotic manipulation tasks. This model, named TecNet, is comprised
of an embedding network and a control network. The embedding network produces a vector
representation of a task given many examples from that task, while the control network
takes an observation and a task representation as input to produce an action. Instead of
computing a task embedding from expert demonstrations, \cite{achille_2019} constructs
them from the Fisher Information Matrix of a pre-trained network. Lastly,
\cite{lan_2019} trains a shared policy for meta-reinforcement learning which is
conditioned on task embeddings. These task embeddings are the outputs of a task encoder
which is trained to output embeddings based on experience from each task.

\section{Multi-Task Benchmarks} \label{benchmarks}
In this section, we give a short overview of commonly used benchmarks in various domains
of multi-task learning, including benchmarks for computer vision, natural language
processing, reinforcement learning, and multi-modal problems. It should be noted that,
while there are a few benchmarks specifically designed for multi-task learning (such as
Taskonomy \cite{zamir_2018} and Meta-World \cite{yu_2019}), these are few and far
between. Most MTL methods are evaluated in multi-task settings which use generic
benchmarks that include supervision for multiple tasks, such as NYU-v2
\cite{silberman_2012}. Lastly, the benchmarks discussed here aren't an exhaustive list,
just a highlight of some of the most commonly used MTL benchmarks. The benchmarks within
each domain are sorted by release date starting with the earliest.

\subsection{Computer Vision Benchmarks}
\begin{itemize}
    \item \textbf{NYU-v2} \cite{silberman_2012} is a dataset of RGB-depth images from
        464 indoor scenes with 1449 densely labeled images and over 400,000 unlabeled
        images. The labeled images are labeled for instance segmentation, semantic
        segmentation, and scene classification, and all images contain depth values for
        each pixel. All images are frames extracted from video sequences.
    \item \textbf{MS-COCO} \cite{lin_2014} contains 328,000 images of natural scenes
        with a total of 2.5 million object instances spanning 91 object types. The
        images contain labels for image classification, semantic segmentation, and
        instance segmentation.
    \item \textbf{CelebA} \cite{liu_2015b} has 200,000 images of celebrity faces, with
        20 images of 10,000 different people. Each image is labeled with 40 face
        attributes and five keypoints, for a total of 8 million facial attribute labels.
    \item \textbf{OmniGlot} \cite{lake_2015} contains images of characters, unlike many
        of the other popular natural image benchmarks. The dataset contains images of
        1623 characters from 50 different alphabets, operating in a low-data regime.
        Omniglot was designed with a focus on few-shot learning and meta-learning in
        image classification and generative modeling.
    \item \textbf{CityScapes} \cite{cordts_2016} is comprised of video frames shot in
        the streets of 50 urban cities. The densely labeled subset of the dataset
        contains 5000 images with pixel-level annotations, while 20000 other images are
        coarsely labeled. The images are labeled for image classification, semantic
        segmentation, and instance segmentation.
    \item \textbf{Taskonomy} \cite{zamir_2018} may be the only large scale computer
        vision dataset specifically intended for research with multi-task learning. The
        dataset consists of 4 million images of indoor scenes from 600 different
        buildings, and each image is annotated for 26 different visual tasks, including
        2D, 2.5D, and 3D tasks.
\end{itemize}

\subsection{Natural Language Processing Benchmarks}
Unless otherwise specified, it can be assumed that the text within a corpus is English.
\begin{itemize}
    \item \textbf{Penn Treebank} \cite{marcus_1993} is a corpus of text consisting of
        4.5 million words. The text is aggregated from multiple sources including
        scientific abstracts, news stories, book chapters, computer manuals, and more,
        and contains Part-of-Speech tags and syntactical structure annotations.
    \item \textbf{OntoNotes 5.0} \cite{weischedel_2013} is a multi-lingual corpus of
        Arabic, English, and Chinese text with 2.9 million words total, labeled for
        syntax and predicate argument structure, coreference resolution, and word sense
        disambiguation. The text sources consist of written news, broadcast news, web
        data, and more.
    \item \textbf{WMT 14} \cite{bojar_2014} is a dataset from the 2014 Workshop on
        Statistical Machine Translation, with parallel corpuses of many language
        pairs, including French-English, German-English, Hindi-English, Russian-English,
        and Czech-English. These corpuses vary in size between 90 million total English
        sentences and 1 million total Hindi sentences.
    \item \textbf{Stanford Natural Language Inference} \cite{bowman_2015} contains
        570,000 sentence pairs, where each pair contains a label describing their
        relationship as either neutral, entailment, or contradiction. The dataset was
        acquired through Amazon Turk, with the instructions displaying a captioned image
        and asking for one alternative caption, one caption that may be correct, and one
        caption that is certainly incorrect.
    \item \textbf{SciTail} \cite{khot_2018} is a textual entailment dataset consisting
        of scientific statements. The corpus was constructed by converting multiple
        choice questions on science exams (and web data) into entailed and non-entailed
        pairs, for a total of 27,000 total examples.
    \item \textbf{GLUE} \cite{wang_2018} consists of nine NLP tasks accompanied with
        data from previously existing NLP corpuses. The benchmark is intended to be used
        to evaluate general language understanding models that can handle all or
        multiple tasks simulataneously. Some tasks are intentionally provided with small
        amounts of training data to encourage information sharing between tasks.
    \item \textbf{decaNLP} \cite{mccann_2018} is a collection of ten NLP tasks which are
        all posed as question answering. This is a new approach to NLP benchmarking:
        instead of the task being specified by explicit constraints on the input/output,
        each task is given to the model with a description in natural language. Each
        example is a 3-tuple of question, context, and answer.
\end{itemize}

\subsection{Reinforcement Learning Benchmarks}
\begin{itemize}
    \item \textbf{Arcade Learning Environment} \cite{bellemare_2013} (or ALE) is a
        diverse collection of hundreds of Atari 2600 games, where observations are given
        to the agent as raw pixels. These games were originally designed to be a
        challenge for the human video game player, so they present a challenge for
        modern RL agents in aspects such as exploration and learning with sparse
        rewards.
    \item \textbf{DeepMind Lab} \cite{beattie_2016} is a 3D first person game platform
        which requires the agent to make actions from raw pixels. DeepMind Lab offers
        the ability to customize environments through the observations, termination
        conditions, reward functions, and more. The 3D nature of the environment makes
        for a challenge not just in strategic decision making but also in perception.
    \item \textbf{Meta-World} \cite{yu_2019} is a collection of robotic manipulation
        tasks designed to encourage research in multi-task learning and meta-learning.
        The collection consists of 50 tasks for a simulated Sawyer robotic arm, each
        task with its own parametric variations, such as goal position. The multi-task
        benchmarks within Meta-World are MT10 and MT50, which consist of simultaneously
        learning 10 and 50 tasks, respectively, while the meta-learning benchmarks are
        ML10 and ML45, which consist of learning on 10 and 45 tasks before being asked
        to quickly adapt to new unseen tasks.
\end{itemize}

\subsection{Multi-Modal Benchmarks}
\begin{itemize}
    \item \textbf{Flickr30K Captions} \cite{young_2014} is a collection of 30,000
        photographs obtained from the image hosting website Flickr, with over 150,000
        corresponding captions.
    \item \textbf{MS-COCO Captions} \cite{chen_2015} contains over 1.5 million captions
        of more than 300,000 photos from the MS-COCO \cite{lin_2014} dataset. These
        captions were collected using Amazon Mechanical Turk.
    \item \textbf{Visual Genome} \cite{krishna_2017} is made of over 100,000 densely
        annotated images with a focus on a grounding connection between visual and
        linguistic concepts. Each image contains over 40 regions which each have their
        own description, 17 (on average) question-answer pairs per image, an average of
        21 object annotations per image, attribute labels per object, and relationship
        annotations between objects.
    \item \textbf{Flickr30K Entities} \cite{plummer_2015} augments the Flickr30K dataset
        with 276,000 annotated bounding boxes and 244,000 coreference chains. The
        coreference chains identify when references to objects in different captions of
        the same image are referring to the same object.
    \item \textbf{GuessWhat?!} \cite{de_2017} is not just a dataset, but a
        dialogue-based guessing game in which a questioner asks an oracle about an
        unknown object pictured in a given image. The paper includes a collection of
        150,000 games played by humans, with 800,000 visual question answer pairs on
        66,000 images. The intention of GuessWhat?! is to introduce a task which bridges
        visual question answering with dialogue.
    \item \textbf{VQA 2.0} \cite{goyal_2017} is a visual question answering dataset
        constructed by balancing the VQA \cite{antol_2015} dataset with a focus on the
        visual aspect of visual question answering. The paper points out that a model
        can reach decent performance on many VQA benchmarks based only on scene
        regularities and the question at hand while ignoring the visual input. VQA 2.0
        is balanced in the sense that every question is accompanied by two images that
        lead to different answers, so that a successful model must pay attention to the
        visual content of a given image.
    \item \textbf{GQA} \cite{hudson_2019} is another visual question answering dataset,
        constructed by leveraging scene graphs to create 22 million questions for their
        corresponding images. The programmatic construction of the dataset allowed for
        each question to be accompanied by a functional program which characterizes the
        question semantics, and for the distribution of answers to be tuned to minimize
        bias.
\end{itemize}

\section{Conclusion} \label{conclusion}
We have presented a review of the field of multi-task learning, covering the three broad
directions of architecture design, optimization techniques, and task relationship
learning. Currently, key techniques for the construction of multi-task neural networks
include shared feature extractors with task-specific decoders, varying parameter sharing
schemes in existing network architectures, sharing and recombination of neural network
modules, learning what to share, and fine-grained parameter sharing. The most prominent
directions within optimization are per-task loss weighting, such as by uncertainty or
learning speed, regularization with $L_2$ and trace norms, gradient modulation and
replacement to avoid conflicting gradients between tasks, and multi-objective
optimization. Finally, several methods have been proposed to learn relationships between
tasks, such as large-scale empirical studies to determine which tasks exhibit positive
learning dynamics when learned simultaneously, comparing representations of networks to
determine task similarity, and learning task embeddings. 

Despite the progress the community has made so far to develop multi-task learning for
deep networks, there is one direction of research that has had less development than
others, and that we have not discussed at all so far: theory. This shouldn't come as a
surprise, given that this is also true of deep learning in general. Still, many
non-neural multi-task learning methods are motivated and justified by strong theory
\cite{baxter_2000, ben_david_2008, zhang_2015, lounici_2009}, but aside from a small
pool of recent work \cite{shui_2019, ndirango_2019, d'eramo_2020, wu_2020, zhang_2020,
bettgenhauser_2020}, there is a lack of theoretical understanding of MTL with deep
neural networks. This is an important area to promote a deeper understanding of the
field as a whole, and we hope to see more development in this direction in the coming
years.

We believe that the development of multi-task learning (and the related fields of
meta-learning, transfer learning, and continuous/lifelong learning) is an important step
towards developing artificial intelligence with more human-like qualities. In order to
build machines that can learn as quickly and robustly as humans, we must create
techniques for learning general underlying concepts which are applicable between tasks
and applying these concepts to new and unfamiliar situations. Building systems that
truly exhibit these qualities will require approaches from many different directions,
likely including many that researchers haven't yet discovered. The field has come a long
way, but continued effort from the research community is needed to fully achieve the
potential of multi-task methods.

\bibliography{references}

\end{document}